\documentclass{SCIS}

\begin{document}

\ArticleType{Review}
\SpecialTopic{Special Topic: Industrial Artificial Intelligence}
\Year{2024}
\Month{February}
\BeginPage{1} 
\EndPage{??}

\title{Computer vision tasks for intelligent aerospace perception: An overview}{Computer vision tasks for intelligent aerospace perception: An overview}
\author[1]{Huilin Chen}{}
\author[1]{Qiyu Sun}{}
\author[2]{Fangfei Li}{lifangfei@ecust.edu.cn}
\author[1]{Yang Tang}{tangtany@gmail.com}


\AuthorCitation{Chen H L, Sun Q Y, Li F F, Tang Y}

\address[1]{School of Information Science and Engineering, East China University of Science and Technology, \\
Shanghai {\rm 200237}, China}
\address[2]{School of Mathematics, East China University of Science and Technology, \\
Shanghai {\rm 200237}, China}


\abstract{
Computer vision tasks are crucial for aerospace missions as they help spacecraft to understand and interpret the space environment, such as estimating position and orientation, reconstructing 3D models, and recognizing objects, which have been extensively studied to successfully carry out the missions. However, traditional methods like Kalman Filtering, Structure from Motion, and Multi-View Stereo are not robust enough to handle harsh conditions, leading to unreliable results. In recent years, deep learning (DL)-based perception technologies have shown great potential and outperformed traditional methods, especially in terms of their robustness to changing environments. To further advance DL-based aerospace perception, various frameworks, datasets, and strategies have been proposed, indicating significant potential for future applications. In this survey, we aim to explore the promising techniques used in perception tasks and emphasize the importance of DL-based aerospace perception. We begin by providing an overview of aerospace perception, including classical space programs developed in recent years, commonly used sensors, and traditional perception methods. Subsequently, we delve into three fundamental perception tasks in aerospace missions: pose estimation, 3D reconstruction, and recognition, as they are basic and crucial for subsequent decision-making and control. Finally, we discuss the limitations and possibilities in current research and provide an outlook on future developments, including the challenges of working with limited datasets, the need for improved algorithms, and the potential benefits of multi-source information fusion.}

\maketitle
 


\begin{multicols}{2}

\section{Introduction}
To enhance human space exploration and extend the operational lifespan of spacecraft, significant efforts have been directed towards prolonging spacecraft longevity. Various space missions, such as debris removal \cite{1,2}, space assembly \cite{3,4}, capture \cite{5,6}, maintenance \cite{7}, and refueling \cite{8}, aim to create a safer space environment for on-orbit spacecraft. Aerospace perception \cite{9,10}, which involves extrapolating the space environment based on sensor input, is essential for the successful execution of these missions. However, the harsh space environment, characterized by extreme temperatures and strong radiation, often leads to spacecraft malfunctions and breakdowns. Consequently, ensuring the durability and stability of spacecraft operations in this complex environment has become an urgent problem to solve \cite{11,12,13,14}. On-orbit service (OOS) for both manned and unmanned spacecraft is a core technology aimed at addressing these challenges, encompassing on-orbit life extension, space debris and defunct spacecraft cleaning, and on-orbit manufacturing. Accurate aerospace perception is a critical prerequisite for the successful execution of these OOS missions.
However, conventional OOS methods that rely on astronauts or ground operators suffer from low efficiency and poor adaptability, primarily due to the limited ability of perception technologies to accurately describe the space environment or the spacecraft's status. The emergence of deep learning has made fully autonomous space vehicles feasible and significantly improved the accuracy and comprehensiveness of perception \cite{15,16,17,18}, thus improving the intelligence level of OOS missions \cite{19,20,21}. 

For vision perception, there are ground- and space-based optical imaging systems. Space-based vision perception is robust in optical imaging and the integrity of images captured during observation is largely immune to interference of the Earth's atmosphere. While ground-based optical imagers are more susceptible to imaging disruptions due to factors such as nighttime operation, atmospheric turbulence, adverse weather conditions, and the specific geographic location, which collectively render them less advantageous than their space-based counterparts. Consequently, this survey focuses on space-based vision perception, which is exclusively concerned with the performance of spacecraft in the space environment. By focusing on space-based systems, we aim to explore the full potential of vision perception unencumbered by terrestrial limitations.

During OOS missions, the service spacecraft and the target first reach a rendezvous and docking distance, after which perception systems on the service spacecraft are employed to perform pose estimation, 3D reconstruction, and recognition tasks. The three tasks are crucial for subsequent decision-making and control, and have been extensively investigated by researchers. Pose estimation is the primary objective of autonomous space missions, where the vision system must accurately measure the relative position and attitude of the service spacecraft (end of a robotic arm) with respect to the target. Different methods are employed for relative pose estimation based on prior information of target models \cite{22}. Therefore, 3D models play a vital role in determining the appropriate pose estimation methods to be used. 3D reconstruction involves recovering the basic shape and silhouette of the target, enabling researchers to study its characteristics \cite{23}. At a semantic level, once the models of these targets are obtained, recognition becomes crucial in classifying them into different types or segmenting useful parts. 

The comprehensive workflow is depicted in Figure 1, which is structured to guide the readers through a multi-tiered approach to space-based perception and mission execution. Figure 1(a) presents an agent-centric view, highlighting the capabilities of individual spacecraft agents. Moving to Figure 1(b), the focus shifts to computer vision tasks critical to aerospace perception. The sequence from left to right outlines the progression from pose estimation to 3D reconstruction and finally to target recognition. Figure 1(c) encapsulates the essence of aerospace missions, presenting four classic examples: robotic manipulation and capture, on-orbit assembly, debris removal, and maintenance and refueling. Figure 1 is designed to introduce the reader from the perspective of individual agents, through the layers of perception, and ultimately to the mission-level objectives, providing a logical and structured understanding of the entire process.

Although traditional perception methods such as Kalman Filtering, Structure from Motion, and Multi-View Stereo still dominate the field, DL-based perception has immense potential as it enables highly intelligent and autonomous space systems. The rise of DL-based perception can be attributed to hardware improvements. Advancements in hardware have facilitated the deployment of sophisticated algorithms on lightweight devices \cite{24}, allowing resource-limited agents to be equipped with multiple devices to tackle challenging tasks. Additionally, robots have made significant progress in perception owing to advancements in sensors. With multi-functional sensors, robots can mimic human abilities to observe, listen, read, and touch. Different sensors provide various data, which robots can exploit for further analysis, enhancing their perception and interaction with the environment \cite{25}. However, there are challenges in the development of aerospace perception, such as insufficient valid data and imperfect results. These challenges can be addressed with the help of advanced AI technologies.

Despite a proliferation of recent surveys within the aerospace field, there remains a notable gap in comprehensive coverage of DL-based technologies and the latest advancements in robotics. Furthermore, while some existing surveys concentrate on particular areas such as robotic manipulation \cite{26}, guidance, navigation and control \cite{27,28,29}, debris removal \cite{1}, among others \cite{30,31}, a scarcity exists in terms of surveys that offer an exhaustive overview of computer vision tasks for aerospace perception, especially those that integrate DL-based, cutting-edge technologies. Against this backdrop, and recognizing the pivotal role of visual perception in On-Orbit Servicing (OOS) missions, this survey endeavors to synthesize the pivotal techniques across three critical computer vision tasks for intelligent aerospace perception: pose estimation, 3D reconstruction, and recognition. We integrate recent Artificial Intelligence (AI)-based technologies to introduce perception tasks within aerospace missions, thereby broadening the scope for researchers to overcome these challenges. Furthermore, we delve into prospective future trajectories, particularly in the context of combining with state-of-the-art advancements in robotics, to offer a forward-looking perspective aimed at augmenting the efficacy of perception tasks. Our synthesis not only fills a critical void in the literature but also paves the way for more sophisticated and effective aerospace perception systems.

The paper is organized as follows. Section~\ref{sec2} provides preliminary knowledge of aerospace perception, including classical space programs, different types of sensors used, and traditional perception methods. Section~\ref{sec3} focuses on recent DL-based pose estimation methods. Section~\ref{sec4} introduces another DL-based perception task, 3D reconstruction, reviewing new reconstruction methods in the field of computer vision and presenting their first application in space settings. Section~\ref{sec5} provides a review of space recognition, which operates at a semantic level. Finally, Section~\ref{sec6} discusses the limitations and possibilities during research, offering an outlook on future development in terms of small datasets, improved algorithms, and multi-source information fusion.
 

\begin{figure*}[htbp]
\centering
    \subfigure[Agents (Spacecraft). These agents, which can be configured as cubesats, are equipped with a suite of sensors tailored to their specific mission requirements, such as optical cameras or active LiDAR systems. They operate collaboratively, leveraging inter-agent communication to perform distributed observation tasks. The advantages of this distributed approach are manifold, including enhanced resilience to space disruptions, agility in task reassignment, improved data accuracy, and an expanded sensing envelope. The agents must contend with the challenge of observing targets that may be in rotation or tumbling motion, which complicates the observation and data acquisition process.]{\includegraphics[width=18.5cm]{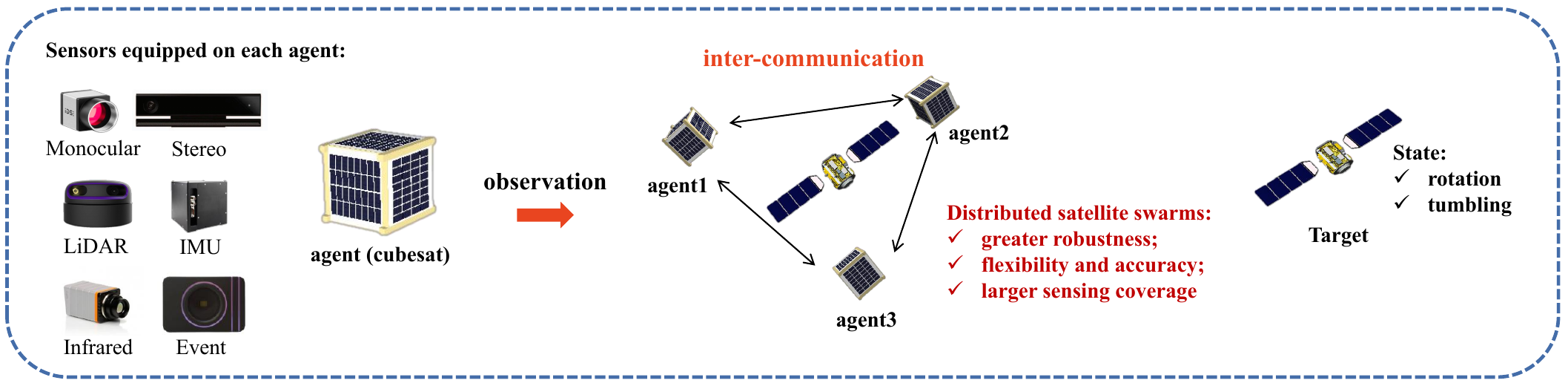}}
    \subfigure[Perception tasks. The sequence from left to right outlines the progression from pose estimation to 3D reconstruction and finally to target recognition. Pose estimation is achieved through two primary methodologies: Kalman Filter (KF) and Machine Learning (ML) techniques. For 3D reconstruction, methods such as Structure from Motion and Multi-View Stereo are employed to ascertain camera positions and generate dense point clouds. Subsequently, advanced algorithms like Neural Radiance Fields (NeRF) are utilized to achieve a detailed reconstruction of the target's geometry and surface features (reprinted from \cite{37}). Recognition tasks are further broken down into object identification (from BUAA-SID 1.0 dataset \cite{38}), which classifies the type of spacecraft, and part recognition, which involves segmenting various components based on their semantic labels.]{\includegraphics[width=18.5cm]{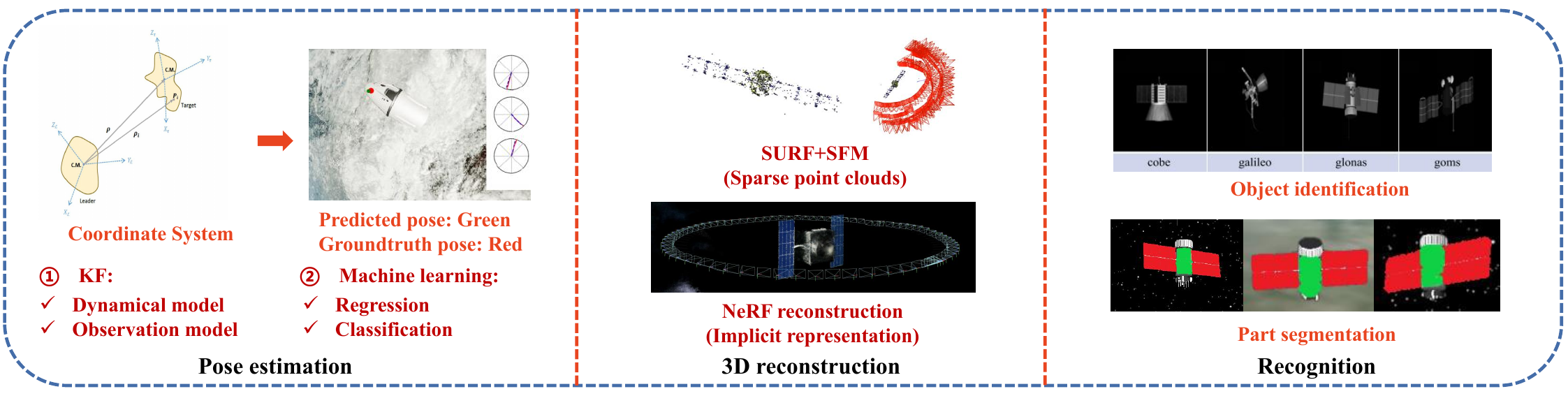}}
    \subfigure[Aerospace missions. Here we present four classic examples: robotic manipulation and capture (reprinted from \cite{26}, copyright © 2021 Papadopoulos, Aghili, Ma and Lampariello); on-orbit assembly (reprinted from \cite{39}, copyright © 2013, NASA); debris removal \cite{14}; maintenance and refueling (reprinted from \cite{40}, copyright © 2021, NASA). The ultimate ambition is the successful accomplishment of these missions.]{\includegraphics[width=19cm]{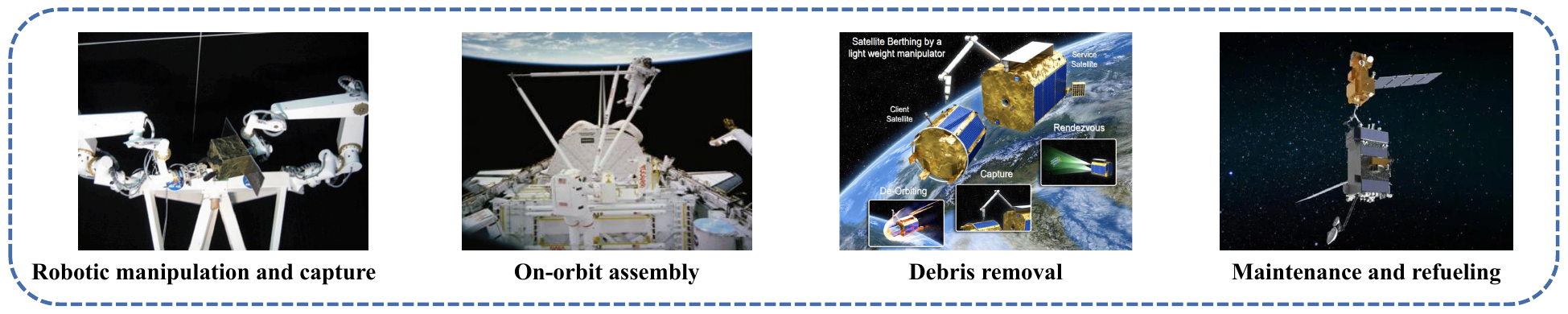}}
    \caption{A comprehensive workflow to guide the readers through a multi-tiered approach to space-based perception and mission execution. (a) A single agent is equipped with different sensors and collaborates with each other to observe the designated target. (b) Through inter-communication, agents can carry out various perception tasks in a distributed manner. (c) Perception is the prerequisite for many aerospace missions. Here we list a series of classical missions.}
    \label{fig: 1}
\end{figure*}

\section{Preliminaries for aerospace perception}\label{sec2}

\begin{figure}[H]
\centering
    \subfigure[DEOS.]{\includegraphics[width=4cm,height=3cm]{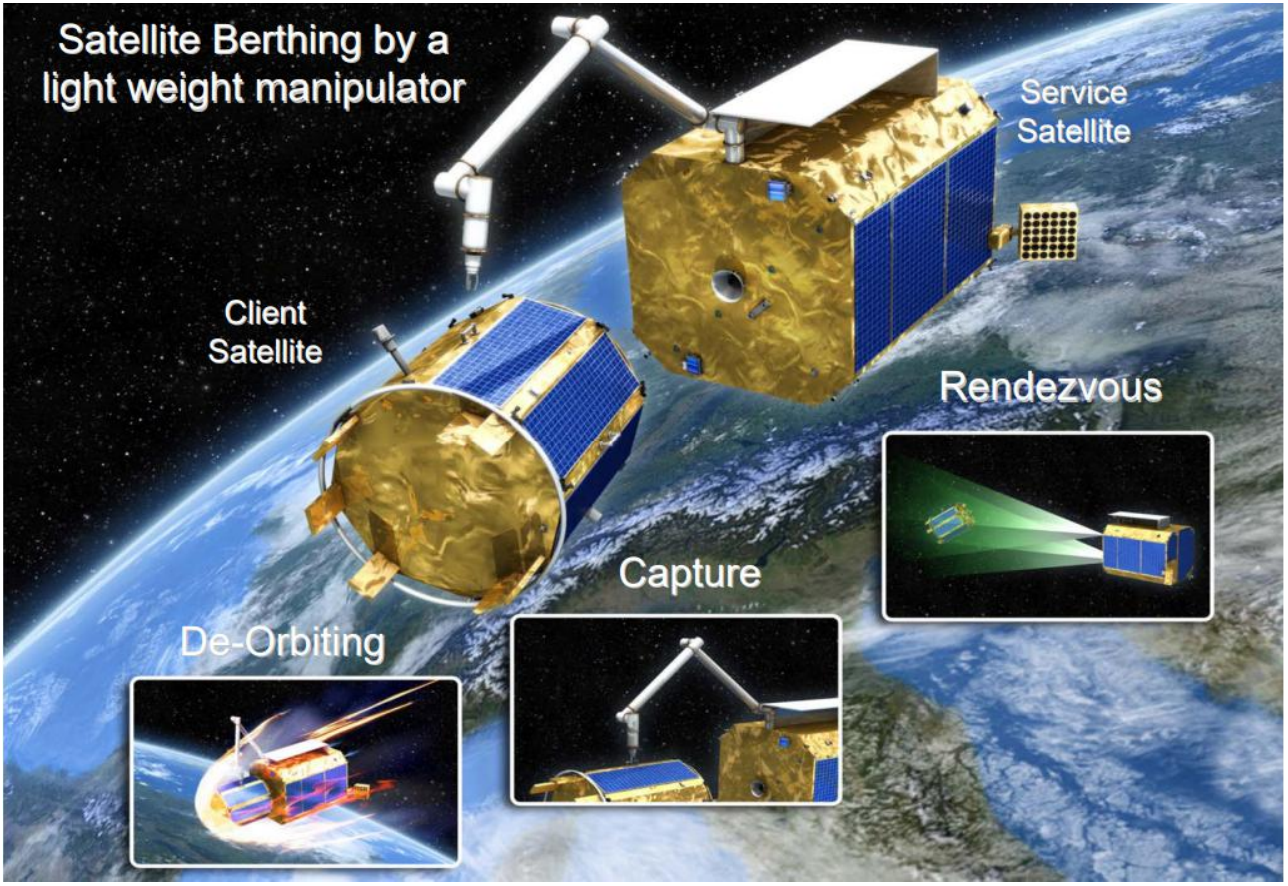}}
    \subfigure[E. Deorbit.]{\includegraphics[width=4cm,height=3.05cm]{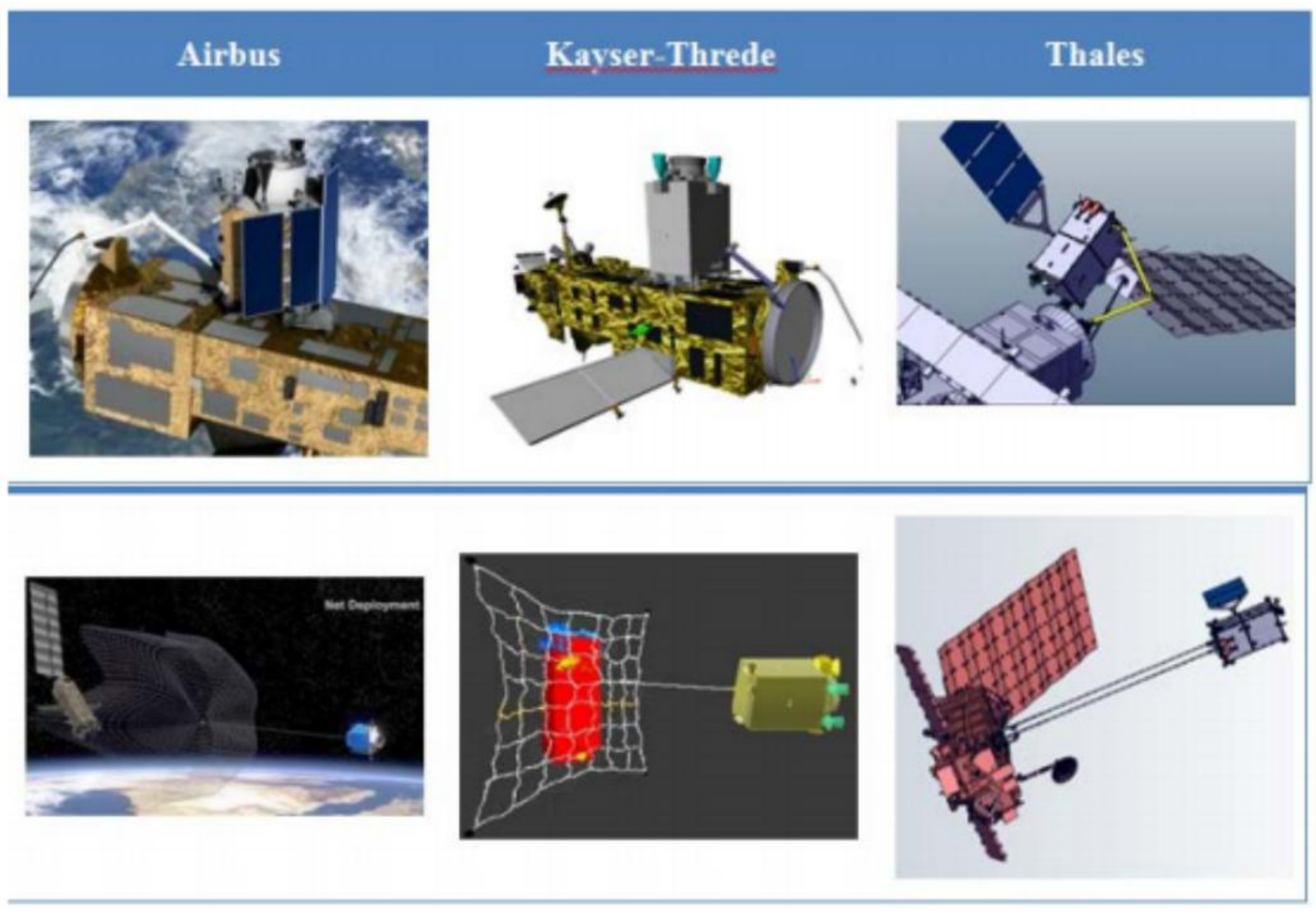}}
    \caption{Example pictures of (a) DEOS demonstration \cite{14} and (b) E. Deorbit demonstration (reprinted from \cite{36}, copyright © ESA/Airbus Space/OHB Systems/Thales Alenia Space).}
\label{fig: 2}
\end{figure}

Many researchers have paid great attention to OOS missions. Numerous task demonstrations have been carried out during the development. In this section, classical space programs from different countries will be introduced firstly. Example pictures of some demonstrations are shown in Figure \ref{fig: 2}. Then a taxonomy of sensors used in perception systems will be discussed. Finally, some traditional perception methods for pose estimation and 3D reconstruction will be introduced.

\subsection{Classical space programs}
Table \ref{tab1} summarizes six classical space programs developed either for technical demonstration or specific space missions. In addition, Table \ref{tab1} also lists their components of visual perception systems, which are especially used in computer vision tasks for aerospace perception.

The XSS-10 micro-satellite demonstration \cite{32}, conducted in 2003 by the United States Air Force Research Laboratory (AFRL), was a pioneering effort aimed at showcasing the capabilities of autonomous navigation, proximity operations, and inspection of Remote Space Objects (RSOs). The XSS-10 was meticulously engineered with a suite of advanced systems, including a miniaturized communication system, streamlined avionics, and a high-resolution integrated camera designed to enable detailed close-up inspections.
The satellite's visual system was a sophisticated dual-camera setup, consisting of a visible star tracker, which was utilized to capture images of celestial bodies for navigational purposes, and a visible imager, tasked with acquiring high-quality photographs of the target RSO. This innovative approach to space-based imaging and navigation underscored the role of XSS-10 as a trailblazer in the field of autonomous space operations.

Front-end Robotics Enabling Near-term Demonstrations (FREND) \cite{22} funded by DARPA was specified at non-cooperative targets in orbit. The robot arm at the end of the spacecraft grasped the target's star and arrow docking ring or other parts to perform rendezvous and capture. The visual system of the FREND arm consisted of three cameras equipped with illumination avionics \cite{33}. When the imaging effect was poor using only one camera, the other two cameras could be paired to form binocular stereo vision. If the target was normally imaged on the three cameras, the measurement accuracy could be improved through information redundancy. 

Smart Orbital Life Extension vehicle (Smart-OLEV) \cite{11} proposed by some European countries, planned to provide life extension for high value commercial satellites and some fleet management services for satellite operators. The whole system included capture tools, telescopic arm, target support bracket, and visual system. The visual system was composed of a close-range stereo camera and an illumination system \cite{12}. When the service spacecraft reached the rendezvous distance, the stereo vision camera captured the image of the target. Then under the guidance of ground operators, Smart-OLEV approached $0.3m$ away from the target.

Deutsche Orbitale Servicing (DEOS) \cite{13} was a demonstration program designed by Germany in 2007. The aim was to verify the services of maintenance, refueling and repair (non-cooperative and tumbling) in a semi-autonomous way from the ground to non-cooperative satellites in Geostationary orbit, and how to clear non-cooperative satellites and orbital debris. In 2011, the visual system carried by the manipulator was stereo vision, composed of a stereo camera with illumination avionics \cite{14}. Subsequently, a monocular camera with illumination replaced the stereo camera and the field of view was about 60°, ensuring that all objects within the range of the clamp can be photographed during the capture process.

Phoenix proposed by DARPA \cite{34} was a demonstration project aiming to leverage manipulators to carry out capture and maintenance of space satellites in GEO. Phoenix had three manipulators, with one flexible robot arm and two FREND robotic arms. The visual system was composed of three monocular cameras with illumination avionics. When the spatial illumination led to unsatisfactory imaging using only one camera, the remaining two cameras could be used to form a stereo vision system to improve the pose measurement of non-cooperative targets. When all cameras worked normally, the accuracy of the pose measurement could be improved through information fusion.

E. Deorbit was part of the European Space Agency's (ESA) Clean Space Initiative launched in 2012 \cite{35}, with the aim of removing massive non-cooperative targets in solar synchronous and polar orbits (800-1000km). There were two visual systems for different situations \cite{36}. The first group was a vision system for manual monitoring, including a black and white camera plus LED lights. The second group was a visual system for automatic operation, including two cameras and two laser emitters, which could constitute two sets of structured light measurement systems.

\begin{table*}
\centering
\caption{\textbf{Classical space programs and their components of visual perception systems.}}
\scalebox{0.9}{
\begin{tabular}{cp{4.5cm}cp{5cm}p{4cm}}
\toprule 
Space program  &  Institution  &  Year  &  Aim  &  Visual perception system\\
\midrule 
XSS-10 \cite{32}  &  American's Air Force Research Laboratory  &  2003  &  demonstrating the capabilities of autonomous navigation, proximity operations and inspection of RSO  & a visible star tracker and a visible imager\\
FREND \cite{33}  &  DARPA  &  2006  &  servicing non-cooperative targets without custom grappling interfaces  &  three cameras equipped with illumination avionics\\
Smart-OLEV \cite{11}  &  SSC (Sweden)/ Kayser-Threde (Germany)/ Sener (Spain)  &  2007  &  providing life extension and other services for high value commercial satellites  &  a close-range stereo camera and an illumination system\\
DEOS \cite{14}  &  Germany  &  2007  &  finding and evaluating procedures and techniques for rendezvous, capture and de-orbiting of an uncontrollable satellite from its operational orbit  &  a monocular camera with illumination avionics\\
Phoenix \cite{34}  &  DARPA  &  2012  &  leveraging manipulators to carry out capture and maintenance of space satellites in GEO  &  three monocular cameras with illumination avionics\\
E. Deorbit \cite{36}  &  ESA  &  2012  &  removing massive non-cooperative targets in solar synchronous and polar orbits (800-1000km)  &  two visual systems: a black and white camera plus LED lights/two cameras and two laser emitters\\
\bottomrule 
\end{tabular}
}
\label{tab1}
\end{table*}

\subsection{An taxonomy of sensors}
In the realm of On-Orbit Servicing (OOS), optical sensors play an indispensable role in the perception of space targets. Given the unique challenges of the space environment, these sensors are tasked not only with accurately measuring the 6D pose of space targets but also with surmounting the effects of target motion, limited observational windows, and constrained resources \cite{41}. 
Drawing from the corpus of OOS research, a variety of sensors are frequently employed, including active sensors such as laser rangefinders, microwave radars, and Light Detection and Ranging (LiDAR) systems, as well as passive sensors like infrared and visible light cameras \cite{42}. However, while laser rangefinders and microwave radars are adept at providing range and angle measurements, they fall short in ascertaining the targets' attitudes. This limitation stems from the difficulty these sensors face in capturing the surface characteristics necessary for such determinations. Furthermore, the resolution of infrared imaging tends to be inadequate at close ranges.
Consequently, LiDAR systems and visible light cameras have emerged as the primary optical sensors capable of independently managing the autonomous relative navigation of non-cooperative spacecraft in close proximity \cite{43}. These advanced sensors offer the necessary high-resolution imaging and detailed surface feature detection, making them ideal for the precise navigation and inspection tasks that are central to successful OOS operations.

\textbf{Active LiDAR}
LiDAR directly obtains point cloud data of the target surface by transmitting and receiving laser beams without complex image processing steps. There are two types of LiDAR, which can be divided into scanning type and array type. Scanning LiDAR obtains one point cloud data at a time, while array LiDAR obtains the whole point cloud at a time. According to the time and phase difference of using the round trip beam, the array type can be further divided into flash LiDAR and Time of Flight (TOF) cameras respectively. TriDAR vision system developed by Neptec \cite{44} is a proximity guidance sensor applied in autonomous rendezvous and docking missions, which integrates autosynchronous layser triangulation technology with TOF ranging in an optic scanning system to perform acquisition and tracking of the targets. Liu \textit{et al.} \cite{45} utilize the flash LiDAR sensor to directly obtain the point cloud data and make an alignment with the model point cloud data to realize pose initial acquisition and tracking of the target.

The advantage of active LiDAR is that it is robust to illumination changes without complex image processing steps, and the degree of distortion can be reduced by obtaining the complete point cloud at one time. However, the disadvantages are also obvious, such as their low resolution and measurement noise. Besides, they have large volume, mass, and energy consumption and generally require more resource allocation.

\textbf{Passive stereo vision}
Stereo vision can provide highly accurate measurement by taking advantage of the mutual information in the images to perform pixelwise matching \cite{46}. Many researches have been conducted based on stereo vision in aerospace missions. Segal and Gurfil \cite{47,48} carry out some terrestrial demonstration experiments with stereo imaging to track resident space objects, showing the advantage of stereo vision to scan large areas and perform relative state estimation. Qian \textit{et al.} \cite{49} develop a stereo vision-based navigation algorithm to calculate the relative pose and motion of non-cooperative targets without any priori information. Tweddle \textit{et al.} \cite{50} employ a stereo vision system called Visual Estimation for Relative Tracking and Inspection of Generic Objects (VERTIGO) to navigate about non-cooperative RSOs, capturing near-simultaneous images of them and then constructing stereo disparity maps to estimate their geometric centers.

Although there are many experiment foundations for stereo vision technology, the drawbacks still exist and are difficult to resolve. In short, as the stereo vision matching process is done for each pixel in the images, the computation time will scale linearly with the increase of resolution. The working distance of stereo vision is also limited to the baseline. 

\textbf{Passive monocular vision}
Considering the sensor limitations of platform satellites, as well as payload, electrical and computing power, monocular cameras have great advantages, especially for small satellites. Compared with LiDARs, monocular cameras have simple structures, low quality, power consumption, and economic cost. Compared with stereo vision systems, monocular cameras have a longer working range and are not limited by the size of the platform on which they are mounted. In recent years, many studies have been devoted to monocular-based experiments, which is an advancement for the application of monocular cameras. Sean Augenstein \cite{51,52} designs a real-time tracking algorithm to implement autonomous rendezvous, inspection and docking with non-cooperative tumbling targets using a single camera. Deng \textit{et al.} \cite{53} propose a non-cooperative target motion estimation method for autonomous navigation based on the matching process of features from a monocular sequence of images.

However, the only limitation of monocular vision is the lack of depth information, which requires additional information as an auxiliary.

\begin{table}[H]
\centering
\caption{\textbf{Comparison of optical sensors. Differences may exist when referring to different literature \cite{54,55}.}}
\scalebox{0.87}{
\begin{tabular}{ccc}
\toprule 
Types of sensors  &  Detection range  &  Measurement accuracy\\
\midrule 
Scanning LiDAR  &  0.5 $\sim$ 2000 m  &  0.1 $\sim$ 5 mm\\
Flash LiDAR  &  0.5 $\sim$ 500 m  &  10 $\sim$ 20 cm\\
TOF cameras  &  2 $\sim$ 250 cm  &  1 $\sim$ 2 cm\\
Stereo cameras  &  \textless 5 m  &  0.04 m\\
Monocular cameras  &  \textless 20 m  &  0.05 m\\
\bottomrule 
\end{tabular}
}
\label{tab2}
\end{table}

Table \ref{tab2} summarizes the characteristics of the optical sensors mentioned above. Each sensor has its limitations, and it is hard to satisfy the relative navigation requirements for non-cooperative targets based only on one sensor. During navigation, it is necessary to choose some sensors to form a measurement system, to expand the measurement capability, and improve the measurement accuracy. Thus, multi-sensor fusion can be a tendency for the development of intelligent perception.

\subsection{Traditional perception methods}
This part is an overview of traditional perception methods for pose estimation and 3D reconstruction. As recognition task is up to a semantic level, which involves the DL process unavoidably, we don't list recognition works here.
\subsubsection{Traditional pose estimation}
The solution of relative position and attitude of space targets using the measurement data from sensors is referred as pose estimation tasks in perception systems. Reliable pose estimation is essential for on-orbit operations, such as debris removal and formation flying. Traditional pose estimation methods can be divided into three categories: model-based methods, feature-based methods and Bayesian-based methods.

Model-based relative pose estimation techniques necessitate a priori knowledge of the target's 3D model. With this foundational information, algorithms can assess the congruence of observed features with the target's 3D structure and subsequently compute the pose estimation through model registration \cite{56}. However, the precise structural attributes of targets are often elusive, which can restrict the applicability of model-based algorithms.
Consequently, an alternative approach that takes advantage of distinct features or intrinsic characteristics to deduce the relative pose is both meaningful and effective. Feature-based methods necessitate the extraction of crafted features, such as corners, edges, and lines, from the 2D imagery of the targets. After extraction, iterative algorithms typically engage in a feature-matching process across successive frames, aiming to minimize the error criterion to ascertain the optimal pose solution.
While commonly utilized natural features, including point, line, and surface features, may lack robustness and the manual selection of features can compromise autonomy, researchers have proposed enhancements. Some have suggested employing rectangular or space circular features to bolster the robustness, computational efficiency, and accuracy of the pose estimation process \cite{57}. These refined features offer a more resilient framework for feature detection and matching, paving the way for more reliable pose estimation outcomes in challenging environments where traditional features falter.

Bayesian-based methods, including the Kalman Filter (KF) \cite{58,59,60} and the Unscented Kalman Filter (UKF) \cite{61}, are probabilistic recursive filtering techniques that excel at estimating and tracking the dynamic pose of objects. They operate by integrating data from various sensors, such as Inertial Measurement Units (IMUs), Global Positioning System (GPS) devices, and vision sensors, to achieve a fused and accurate representation of the object's pose.
These filtering techniques employ measured values to refine and enhance the pose estimation through two critical stages: the prediction step and the update step. The prediction step forecasts the state transition, while the update step corrects the prediction using new measurements, thus mitigating the impact of noise and errors. In the context of a classical KF model, dynamic equations and measurement equations are formulated to separately articulate the state transition process and the sensor measurement process.
Compared with model-based and feature-based methods, Bayesian-based methods are particularly well-suited for pose estimation in dynamic systems. They are adept at managing pose variations over continuous time intervals and are capable of providing real-time pose estimate updates, making them highly adaptable to environments characterized by constant change.
However, the effectiveness of these methods is contingent upon the availability of accurate initial states. Additionally, it is important to note that the computational complexity escalates significantly with an increasing number of target features integrated into the filter. Despite this, Bayesian-based methods remain a powerful tool for pose estimation tasks, particularly in scenarios involving dynamic and complex systems where the need for real-time updates is paramount.

Table \ref{tab3} lists some pose estimation works, a priori information of their observed targets, and sensors used. Different methods choose distinct features to satisfy their task demands and different priori knowledge of space targets also gives rise to the emergence of different solving ways.

According to the sensors used, different methods will be selected to solve the relative pose. For example, LiDAR can obtain dense point clouds and realize 3D reconstruction of the targets. Therefore, based on ICP (Iterative closet point) algorithm, the pose of non-cooperative targets can be measured through alignment of point clouds \cite{62,63}. However, LiDAR is power-consuming, heavy, and expensive, which is not an ideal and economical solution. There has also been some research using stereo cameras, which is a substitute of LiDAR and is often used in pose estimation tasks for non-cooperative space targets. Stereo cameras can recover depth information through the triangulation method, which is a simulation of human-eye imaging. However, stereo-based methods have short working ranges due to the limited camera baseline. In contrast to LiDAR and stereo cameras, monocular-based pose estimation ensures lower mass and power requirements, therefore being a potential sensor candidate for real-time pose estimation.

\begin{table*}\small
\centering
\caption{\textbf{Traditional perception works.} The left part of this table are some traditional pose estimation works and the right part are traditional 3D reconstruction works. A priori information indicates whether prioir information of the target is used. Sensors: mono. represents monocular cameras, st. represents stereo cameras, dep. represents depth cameras, l.r. represents laser rangefinder and TOF means TOF cameras.}
\begin{tabular}{lccc|lccc}
\toprule 
\multicolumn{4}{c|}{Pose estimation} & \multicolumn{4}{c}{3D reconstruction}\\
\midrule 
\makecell{Literature} & Year & \makecell{A priori information} & Sensors & \makecell{Literature} & Year & \makecell{A priori information} & Sensors\\
\midrule 
Cao et al. \cite{65} & 2015 & \ding{52} & mono. & He \cite{76} & 2017 & \ding{55} & ToF\\
Opromolla et al. \cite{66} & 2015 & \ding{55} & LiDAR & Li et al. \cite{77} & 2017 & \ding{55} & mono.\\
Liu et al. \cite{34} & 2016 & \ding{52} & LiDAR & Dziura et al. \cite{78} & 2017 & \ding{55} & st.\\
Wang et al. \cite{67} & 2016 & \ding{55} & st. & Zhang et al. \cite{79} & 2017 &\ding{55} & mono.\\
Oumer et al. \cite{68} & 2017 & \ding{52} & st. & Ma \cite{64} & 2018 & \ding{55} & LiDAR\\
Shtark et al. \cite{69} & 2017 & \ding{55} & st. & Stacey et al. \cite{80} & 2018 & \ding{55} & mono.\\
Dor et al. \cite{70} & 2018 & \ding{55}& st. & Dor et al. \cite{81} & 2018 & \ding{52} & mono.\\
Sharma et al. \cite{71} & 2019 & \ding{55} & mono. & Wong et al. \cite{82} & 2018 &\ding{55} & mono.\\
Mu et al. \cite{72} & 2020 & \ding{55} & st. & Chen et al. \cite{83} & 2021 & \ding{55} & mono.\\
Ge et al. \cite{73} & 2020 & \ding{55} & st. & Hu et al. \cite{84} & 2023 &\ding{55} & radar\\
Peng et al. \cite{74} & 2020 & \ding{55} & st. & Zeng et al. \cite{85} & 2023 &\ding{52} & dep.\\
Liu et al. \cite{75} & 2022 & \ding{55} & mono. & Dennison et al. \cite{86} & 2023 &\ding{55} & st.\\
\bottomrule 
\end{tabular}
\label{tab3}
\end{table*}

\subsubsection{Traditional 3D reconstruction}
Traditional 3D reconstruction methodologies are often categorized based on the types of sensors utilized on spacecraft. Active sensors, such as Time-of-Flight (TOF) cameras \cite{64} and LiDAR systems \cite{64}, are used to scan the surfaces of non-cooperative targets. These sensors facilitate the reconstruction process by generating depth information and point-cloud data, which are then used to construct a 3D model of the target.
However, it is important to acknowledge that scanning LiDAR systems can be particularly sensitive to variations in lighting conditions and the motion of the target. These factors have the potential to significantly affect the quality of the reconstruction and may even lead to the distortion of the collected point cloud data. Additionally, Flash LiDAR systems are constrained by their range, necessitating greater energy expenditure to sufficiently illuminate the entire field-of-view (FOV) at extended distances.
Table \ref{tab3}, as referenced, provides an overview of various traditional 3D reconstruction efforts. The table highlights that many of these methods operate without any prior knowledge of the target. In this context, monocular cameras have garnered increasing interest due to their ability to generate depth estimates from single images, despite the inherent challenges associated with the lack of stereo information.

Passive sensors such as monocular \cite{64}, stereo \cite{78} and multi cameras \cite{80} firstly obtain the image sequences, then achieve 3D reconstruction based on the multi-view geometry principle. During the reconstruction process, feature extraction and feature matching are key steps. Special features are detected and extracted, and then point clouds can be obtained through triangulation. However, due to the special characteristics of space (no background texture, institutional symmetry, and texture repetition), more matching errors may be produced. 

Structure from Motion (SfM) is a traditional reconstruction algorithm which can obtain inertial parameters of cameras, relative poses and recover sparse point clouds of the whole scene. It is a typical algorithm of passive vision reconstruction. When images are taken over short time intervals between cameras, it will be easier to track feature points and make correspondences \cite{87}. Then, initial poses of cameras can be solved and bundle adjustments can be applied to optimize the poses. Finally, the scene can be reconstructed through inertial parameters of cameras, poses and feature points \cite{88}. Various feature points can be detected such as SIFT (Scale Invariant Feature Transform), SURF (Speeded Up Robust Features), Harris, Fast and ORB features. For natural targets such as asteroids, SIFT keypoints are often used to pick out the various boulders, craters and textures on the surface \cite{89,90}. For man-made targets with few blob-like features, such as spacecraft, ORB features are more favorable, as they tend to detect straight lines, smooth surfaces, and corners \cite{81,91}. 

However, SfM can only recover sparse points, leading to non-ideal reconstruction results. Therefore, multiview stereo (MVS) is an improvement for the SfM algorithm and can recover more dense point clouds. MVS computes the 3D information of most pixels through stereo matching, which can acquire more complete 3D information than sparse point clouds \cite{92}. Chen \textit{et al.} \cite{83} propose CMVS (clustering multiview stereo) to cluster collected images and PMVS (patch-based multi-view stereo) to generate dense point clouds, minimizing the reconstruction errors at the same time. Dennison \textit{et al.} \cite{86} perform a comprehensive assessment of the sensitivity of vision-based 3D reconstruction to different design parameters, suggesting how to optimize a rendezvous mission for fast reconstruction of arbitrary targets.

Currently, many sensor fusion techniques have been proposed to fully exploit different advantages of sensors. For example, depth sensors can be combined with RGB cameras to collect multidimensional data with high resolution. Based on RGB-D cameras, 3D reconstruction can be classified into two types: KFusion \cite{93}, Kintinuous \cite{94} and ElasticFusion \cite{95} for static scenes reconstruction, DynamicFusion \cite{96} for dynamic scenes reconstruction.

\subsubsection{Limitations of traditional perception methods}
Traditional perception methods, constrained by algorithmic and hardware limitations, often yield results with limited precision and slow inference speeds. These methods also struggle to maintain robustness under challenging conditions such as fluctuating illumination, motion blur from rapidly spinning targets, and communication lags. The dynamic nature of space environments can significantly impair the effectiveness of these conventional approaches.
In contrast, with cameras emerging as the sensors of choice for On-Orbit Servicing (OOS) missions due to their compact and lightweight design, vision-based Deep Learning (DL) methods present themselves as compelling alternatives. DL-based techniques offer improved robustness, particularly in the initialization phase of attitude determination, which is a common challenge for traditional methods. Furthermore, the development of sophisticated DL algorithms has the potential to simplify computational processes, reducing complexity and the reliance on intricate dynamic models.
DL methods are also capable of providing supplementary information that can be integrated with navigation filters and other systems, enhancing overall performance. This survey underscores the significant advancements in DL methods within the field of aerospace perception, demonstrating their potential to revolutionize future applications. Using the power of DL, aerospace perception systems can achieve higher accuracy, faster processing times, and improved resilience in the face of environmental uncertainties, paving the way for more reliable and efficient OOS missions.

\section{DL-based pose estimation}\label{sec3}
DL models have risen to prominence in recent years, demonstrating exceptional performance on various computer vision tasks, particularly those involving autonomous or intelligent systems. Within the aerospace industry, pose estimation that uses deep learning has advanced, progressing from conceptual development to practical application.
Given the critical role of pose estimation in the perception process, and considering its prominence as a focal point of research within the DL community, this section will provide an in-depth examination of the subject, emphasizing the significance of the pose estimation task in aerospace perception.
To structure our discussion, we will begin by presenting an overview of the datasets and evaluation metrics commonly utilized in DL-based pose estimation research. This foundation will be followed by an exploration of seminal works in the field, detailing the methodologies and contributions of each. Our aim is not only to outline the current landscape of pose estimation but also to highlight the transformative impact of DL on this pivotal aspect of aerospace perception.

\subsection{Datasets}
It is extremely challenging to train and validate space-borne DL models due to the impracticality of acquiring a large-scale labeled dataset of images of the intended target in the space environment. For a single target, existing space-borne images lack diversity in terms of quantity, pose distribution and variability of environmental factors. Moreover, since they lack accurate pose labels, evaluations are forced to rely on hand-labeled pose annotations. In order to address these problems, some organizations have proposed their datasets constructed for spacecraft pose estimation and space-borne computer vision problems. 

\begin{figure}[H]
\centering
    \includegraphics[width=8cm,height=5cm]{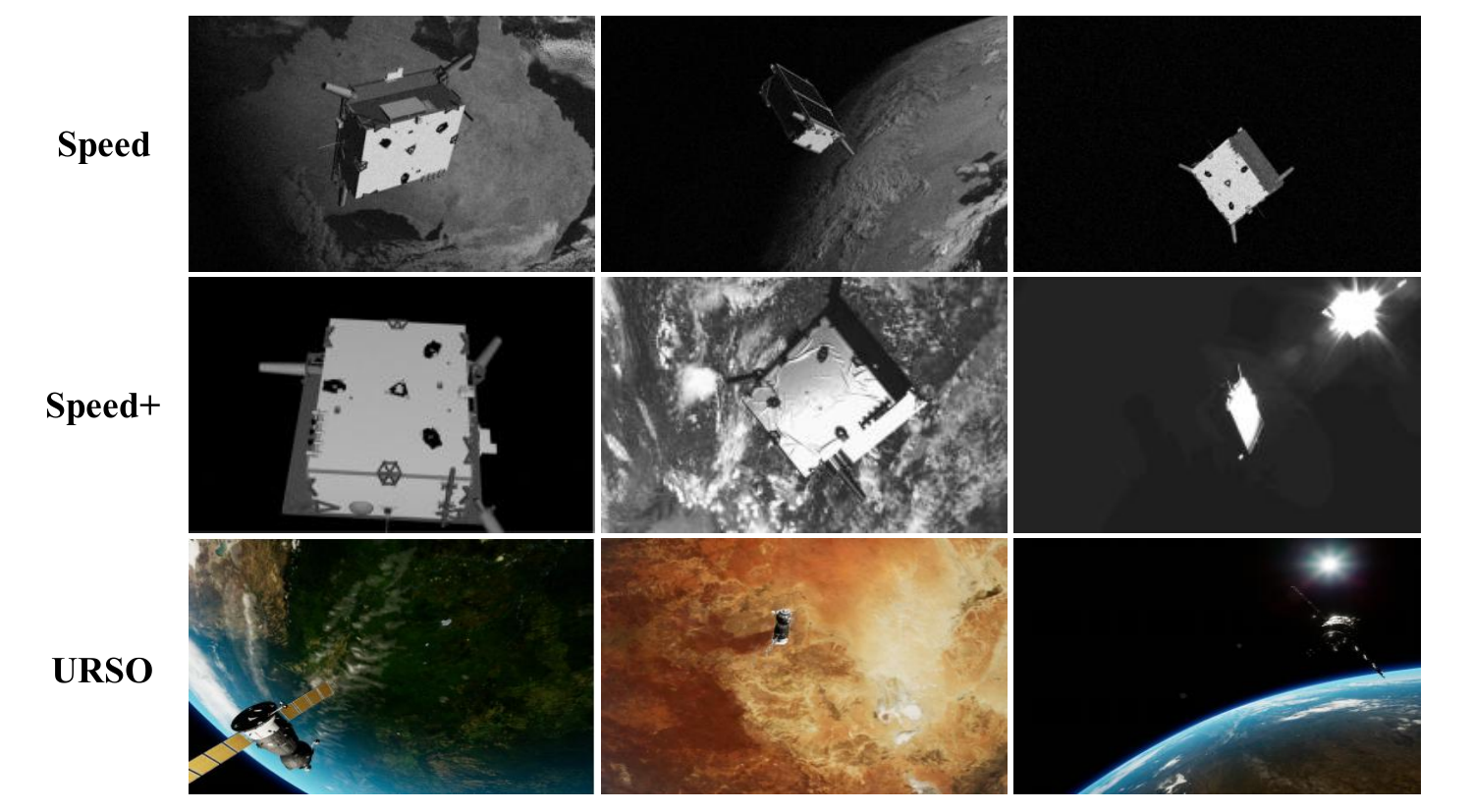}
    \caption{Three datasets for DL-based pose estimation. SPEED: copyright © European Space Agency 2021, SPEED+: copyright © Stanford University, URSO: copyright © Pedro F. Proença.}
\label{fig: 3}
\end{figure}

\textbf{Spacecraft Pose Estimation Dataset (SPEED)}
It is a dataset used for the Satellite Pose Estimation Challenge (SPEC), which was organized by the Space Rendezvous Laboratory (SLAB) at Stanford University and the Advanced Concepts Team (ACT) of the European Space Agency (ESA). SPEED represents the first publicly available machine learning dataset for spacecraft pose estimation \cite{97}. The images of the Tango spacecraft from the PRISMA mission can be classified into two types, referred to as synthetic and real images. Synthetic images were created using the OpenGL-based Optical Stimulator (OS) camera emulator software while the real images were created using a realistic satellite mockup and the Testbed for Rendezvous and Optical Navigation (TRON) facility of SLAB \cite{98,99}. Example images are shown in Figure \ref{fig: 3}. The ground-truth pose labels of training images consist of a translation vector and a unit quaternion describing the relative orientation of the Tango spacecraft with respect to the camera. More details on the dataset can be found in \cite{97}.

\begin{table*}
\centering
\caption{\textbf{Datasets specified at different visual pecerception tasks.}}
\scalebox{0.9}{
\begin{tabular}{ccccc}
\toprule 
Datasets  &  Organizations  &  Year  &  Types of images (numbers)  &  Tasks\\
\midrule 
Speed  &  SLAB $\&$ ACT  &  2019  &  synthetic/15000, real/300  &  Pose estimation\\
Speed+  &  SLAB $\&$ ACT  &  2021  &  synthetic/60000, real/9531  &  Pose estimation\\
URSO  &  Pedro F. Proença $\&$ Yang Gao  &  2019  &  synthetic/5000  &  Pose estimation\\
\midrule 
SatelliteDataset  &  Hoang \textit{et al.}  &  2021  &  synthetic and real/3117  &  Spacecraft detection and segmentation\\
SPARK  &  SnT $\&$ ICIP  &  2021  &  synthetic/150000  &  Spacecraft recognition and detection\\
\bottomrule 
\end{tabular}
}
\label{tab4}
\end{table*}

\textbf{Next-Generation Dataset for Spacecraft Pose Estimation (SPEED+)}
Although there are 300 real images used as a evaluation test set for ML models in SPEED, restricted pose and illumination configurations tend to fail to perform a comprehensive analysis of model robustness across domain gaps. Considering such limitations, SPEED+ \cite{100} comprises more labeled synthetic images for training and adds two simulated HIL domains of unlabeled HIL images for testing. The two domains are referred to as different sources of illumination: lightbox and sunlamp. Example images of SPEED+ are shown in Figure \ref{fig: 3}.

\textbf{Dataset captured using a simulator built on Unreal Engine 4 (URSO)}
A visual simulator built on Unreal Engine 4, named URSO, is proposed in \cite{101}. This simulator can render photorealistic images and depth masks of commonly used spacecrafts orbiting the Earth. The datasets obtained from URSO in \cite{101} are composed of synthetic and spaceborne images of the Soyuz and Dragon spacecraft, one dataset for the dragon spacecraft and two datasets for the Soyuz model with different operating ranges. The three datasets all contain 5000 images, 10$\%$ used for testing and 10$\%$ used for validating, while labels for spaceborne images are missing. Example images of URSO are shown in Figure \ref{fig: 3}.

\subsection{Metrics}
\begin{figure}[H]
\centering
    \includegraphics[width=5cm,height=5cm]{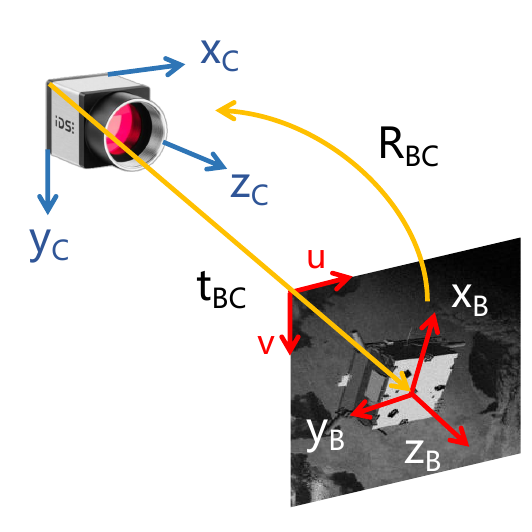}
    \caption{A description for the reference frames to compute the errors. Spacecraft body reference frame ($\mathcal{B}$), camera reference frame ($\mathcal{C}$), relative position ($t_\mathcal{BC}$), relative orientation ($R_\mathcal{BC}$).}
\label{fig: 4}
\end{figure}

The metrics come from the standard competition named SPEC, which is a faithful reflection of the underlying scientific problems. Position ($e_t$) and orientation ($e_q$) errors are evaluated separately. A graphical description of the relevant reference frames is shown in Figure \ref{fig: 4}.

For convenience, we define the spacecraft body reference frame as $\mathcal{B}$ and camera reference as $\mathcal{C}$.
For the position error, $e_t$ is defined as
\begin{equation}
    e_t = \|t_\mathcal{BC}-\hat{t}_\mathcal{BC}\|_2,
\end{equation}

where $t_\mathcal{BC}$ and $\hat{t}_\mathcal{BC}$ represent the ground truth and the estimated position vector from the origin of $\mathcal{C}$ to the origin of $\mathcal{B}$ respectively. When the target satellite is closer, the position errors should be penalized more heavily, thus the normalized position error $\overline{e}_t$ is introduced as
\begin{equation}
    \overline{e}_t = \frac{e_t}{\|t_\mathcal{BC}\|_2},
\end{equation}

For the rotation error, $e_q$ is defined as
\begin{equation}
    e_q = 2\arccos(|\langle\hat{q},q\rangle|),
\end{equation}
where $e_q$ represents the orientation error, $\langle\hat{q},q\rangle$ denotes the inner product of the two quaternions. $e_q$ is calculated as the angular distance between $q=q(R_\mathcal{BC})$ (ground truth) and $\hat{q}=q(\hat{R}_\mathcal{{BC}})$ (estimated), aligning the target body frame $\mathcal{B}$ with the camera reference frame $\mathcal{C}$.

For one single image, the pose error $e_{pose}$ is a summation of the normalized position error $e_t$ and rotation error $e_q$,
\begin{equation}
    e_{pose} = \overline{e}_t+e_q,
\end{equation}

Finally, the average pose error across all test images is used to measure the accuracy of the prediction results. Assuming that there are N images in the test set, the overall pose error E can be calculated as
\begin{equation}
    E=\frac{1}{N}\sum_{i=1}^{N}{e^{(i)}}_{pose},
\end{equation}

\subsection{Fundamental works}
Previous monocular methods for spacecraft tracking and pose estimation have predominantly relied on model-based solutions, which involve aligning a wireframe model to the edge image of the target. However, these traditional approaches often overlook the intricate correlations between features and the rich set of characteristics that define objects, extending beyond mere edges and geometric primitives.
Convolutional Neural Networks (CNNs), on the other hand, have the capacity to discern a more comprehensive array of features from imagery, showcasing robustness even in the presence of complex backgrounds. This ability has not only positioned CNNs as a formidable tool in the computer vision field but also signaled their great potential for space-borne pose estimation tasks.
In recent years, there has been a surge in the development of CNN-based methods specifically tailored for application in the domain of spacecraft pose estimation. These methods leverage the power of deep learning to enhance the accuracy and reliability of pose estimation, offering promising alternatives to the model-based techniques that preceded them. The adoption of CNNs in this context marks a significant step forward in the evolution of spacecraft tracking and pose estimation methodologies.

A CNN-based Spacecraft Pose Network (SPN) was first introduced in \cite{102}, which just required a grayscale image as input and output the final pose of the spacecraft relative to the monocular camera. It incorporates a state-of-the-art detection algorithm to determine the bounding area of the target in the first branch of the network. The other two branches are used to predict the position and attitude based on classification. The results are generally satisfactory trained and tested on SPEED dataset. However, when faced with too large or too small images of spacecraft, the SPN method tends to perform poorly. It also lacks robustness to data distributions as the network is trained soly on synthetic images.
To address the aforementioned challenges, a novel CNN architecture which merges state-of-the-art detection and pose estimation pipelines with MobileNet architecture, has been introduced in \cite{103} for fast and accurate pose estimation. By regressing the 2D locations of spacecrafts’ surface keypoints and directly using publicly available PnP solvers, the pose estimation can be carried out computationally. Besides, introducing the Neural Style Transfer (NST) technique to randomize the object texture leads the CNN to learn global shape instead of focusing on local texture of the target. However, this process is divided into four steps and not end-to-end.
In 2020, an end-to-end framework for satellite pose estimation has been proposed in \cite{101}, which adopts the ResNet architectures with pre-trained weights as the network backbone, considering its fewer pooling layers and a trade-off between accuracy and complexity.
In the Kelvin competition, the approach from \cite{104} demonstrated superior accuracy and won the first place. They firstly recover the 3D coordinates of arbitrarily chosen landmark points via multi-view triangulation, then a deep network to predict the position of these landmark points in the input image will be trained. Finally, given predicted 2D-3D correspondences, a robust nonlinear optimization to compute the pose estimates can be performed.
Apart from multi-view triangulation to recover the landmark points, \cite{105} takes advantage of 3D structure as a priori information, which proposes a novel CNN combined with 3D model in the form of point clouds. They assume that the uncooperative target satellite is a known object of which the basic geometry is available.
To handle with varying lighting conditions, low resolution, and limited amount of data, \cite{106} proposes a dense residual U-Net network to effectively extract orbiting spacecraft features from images, delivering strong performance gains on pose estimation.

Despite the promise of learning-based pose estimation methods, several challenges must be addressed to ensure their deployability in practical applications. Firstly, the robustness of these methods hinges on their ability to generalize beyond synthetic imagery to real-world space imagery. This requires training and testing networks on a diverse and representative set of data that accurately reflects the conditions and variations encountered in space.
Secondly, enhancing the performance of Convolutional Neural Networks (CNNs) necessitates access to a significantly larger dataset. Ideally, this dataset should be scalable and encompass a wide range of target spacecraft to accommodate the variability in shapes, sizes, and features that pose estimation must contend with.
Additionally, many current CNN architectures presuppose some knowledge of the target's shape and geometry, which may not always be readily available or accurate in real missions. This limitation can hinder the applicability of these methods in scenarios where a priori information is limited or uncertain.
Lastly, for on-orbit servicing missions, real-time capability is imperative. Any proposed CNN must be pre-evaluated on spacecraft hardware to confirm its suitability and ensure that it meets the stringent latency requirements of such operations.
In summary, future research endeavors in pose estimation must confront these challenges head-on. By doing so, the field can fully harness the potential of learning-based methods, paving the way for more accurate, reliable, and efficient spacecraft tracking and pose estimation in real-world space missions. Addressing these issues will be crucial for the translation of these methods from theoretical constructs to practical tools that can significantly enhance the capabilities of aerospace operations.


\section{DL-based 3D reconstruction}\label{sec4}
3D reconstruction is a key technology in on-orbit service missions, assisting space tasks like safely approaching, orbiting, capturing and docking with the target of interest. When the observed target is non-cooperative, 3D reconstruction is particularly important and indispensable for pose estimation. Some steps of 3D reconstruction even become an integral part of pose estimation process, such as detecting, identifying and tracking features on the targets \cite{107,108}. Up to now, there have already been many researches on 3D reconstruction of space targets.

Traditional 3D reconstruction methods rely on handcrafted image features and matching metrics, which are prone to incomplete structures caused by hard feature detection and inaccurate feature matching. Recently, DL-based methods are introduced for more robust feature matching. Initially, attempts are made for two-view stereo matching by substituting hand-crafted similarity metrics \cite{109,110} or engineered regularizations \cite{111,112} with the learned ones. However, for multi-view scenarios, just merging all pairwise reconstructions through stereo matching will fail to exploit multi-view information and lead to inaccurate results. Many works realize this problem and try to apply CNN to tackle it. SurfaceNet \cite{113} is the first learning-based pipeline for MVS problems, which can automatically learn the photo-consistency and geometry relations of the surface structure. The Learned Stereo Machine (LSM) \cite{114} directly leverages the underlying 3D geometry based on the differentiable projection/unprojection, enabling reconstruction from fewer images or even one single image. As the two methods above exploit volumetric representations that are restricted by memory consumption, their networks can hardly scale up. To avoid this problem, MVSNet \cite{115} focuses on computing one depth map for the reference image at each time, which can adaptively reconstruct large scenes. With the promising future of 3D technology, many improved works based on MVSNet \cite{116,117,118} have emerged these years.


Apart from explicit 3D representations such as point clouds and voxels, implicit representations \cite{119,120} have been investigated in recent years, which represent continuous 3D shapes by mapping $\textit{xyz}$ coordinates to signed distance functions or occupancy fields. However, they have been limited to simple shapes, oversmoothed renderings until Neural Radiance Field (NeRF) \cite{121} appears. NeRF encode 5D radiance fields, which refer to 3D location $\bm{x}=\textit{(x,y,z)}$ and 2D viewing direction \textit{($\theta$,$\phi$)}, to represent a continuous scene and output emitted color $\bm{c}=\textit{(r,g,b)}$ and volume density \textit{$\sigma$}. With many NeRF-based works springing up, NeRF is always under optimization \cite{122,123,124} and becomes an increasingly popular technique in real applications, even in dynamic \cite{125,126} and dark environments \cite{127,128}. Many works also consider the characteristics of observed objects such as transient or specular ones.

\begin{figure}[H]
\centering
    \subfigure[Camera visualization.]{\includegraphics[width=8cm,height=2.3cm]{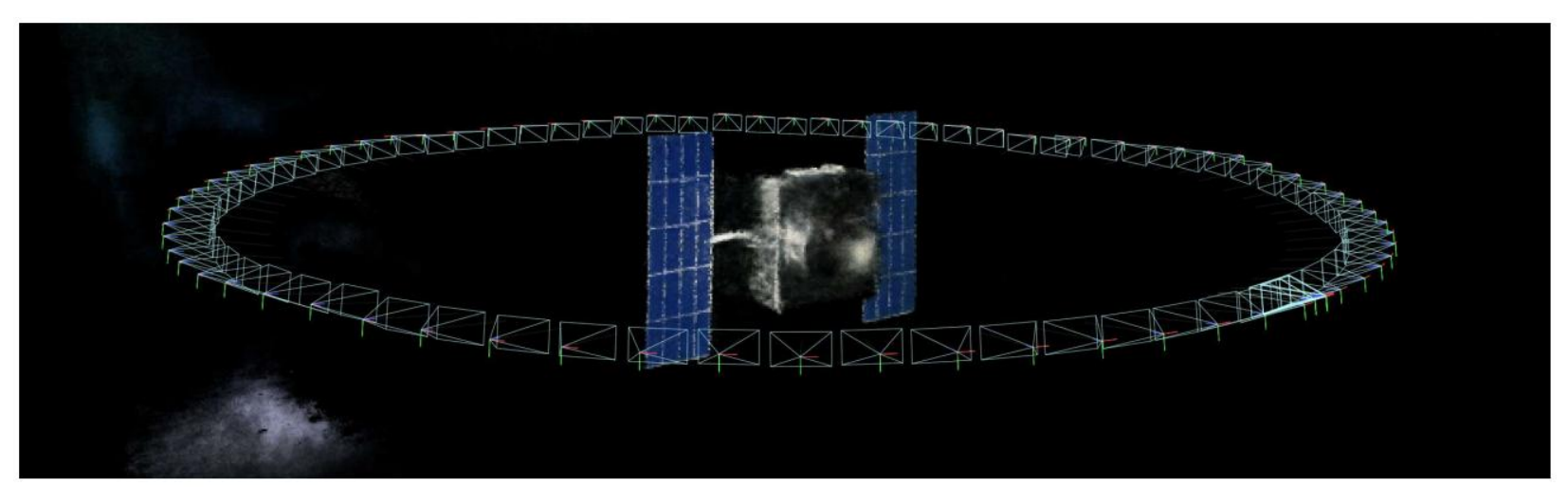}}
    \subfigure[Reconstruction results.]{\includegraphics[width=8cm,height=3.5cm]{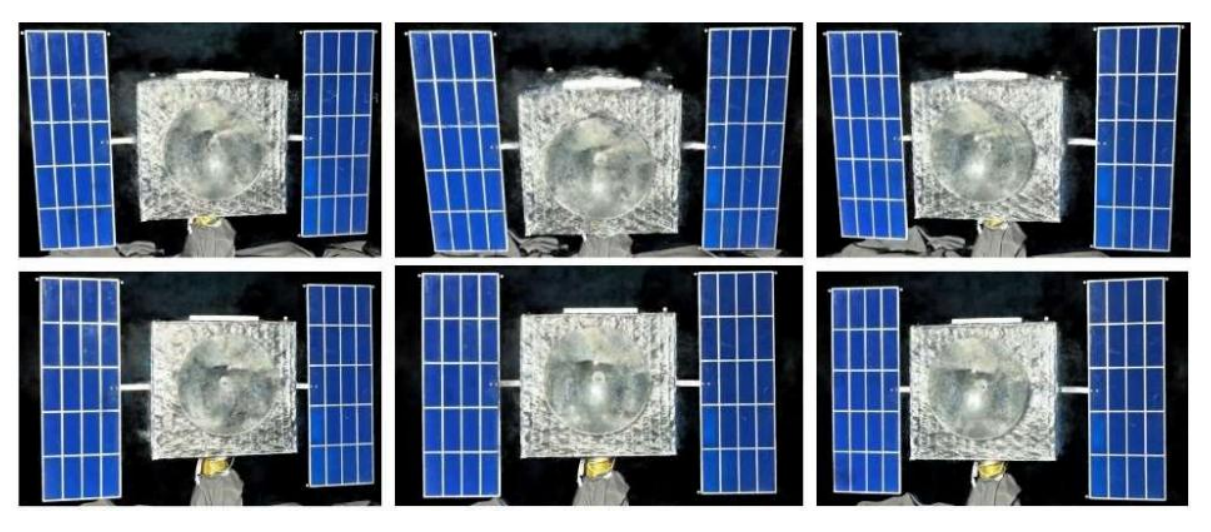}}
    \caption{Experiment images from \cite{37}.}
\label{fig: 5}
\end{figure}

In fully autonomous advanced mission concepts, a high-fidelity 3D model of the target RSO is required to have a better knowledge of the unfamiliar target and reduce the risk of collision. As monocular cameras have the advantage of being light-weight, small in size, low in power consumption, and low in cost, many aerospace missions focus on 3D reconstruction of the targets based on a set of monocular images before approaching them. DL-based view synthesis algorithms have been proved by ESA \cite{129} to have the ability to infer the 3D geometry of an RSO. With NeRF showing great success in novel view synthesis tasks, it has a huge potential in reconstructing the geometry of space targets. The ESA paper \cite{129} compares and evaluates the ability of NeRF \cite{121} and GRAF \cite{130} to render novel views and infer the 3D shapes of two different spacecraft, showing that NeRF can render more accurate images with specular materials while GRAF can generate novel views with precise details. Caruso \textit{et al.} \cite{37} test the potential of NeRF, instant neural graphics primitive (Instant-NGP) \cite{122}, D-NeRF \cite{125} under space-like conditions. Some experiment images are shown in Figure \ref{fig: 5}.

\section{DL-based Recognition}\label{sec5}
Recognition technology is a critical component in space missions, encompassing the precise detection, localization, and segmentation of objects within spaceborne imagery. This process is not only a fundamental task in its own right, but also serves as an essential precursor to vision-based pose estimation, which is vital for the successful execution of various space operations.
In the realm of DL-based recognition for spacecraft, a multitude of studies and approaches have been developed. These methods have made significant strides in improving the accuracy and reliability of spacecraft recognition from spaceborne images. In this section, we aim to provide a comprehensive review of some of the foundational works that have laid the groundwork in this area. We delve into the methodologies and innovations that have propelled the field forward while also acknowledging the current limitations that persist.

\begin{figure}[H]
\centering
    \subfigure[Examples of collected images \cite{132}.]{\includegraphics[width=7.3cm,height=2.8cm]{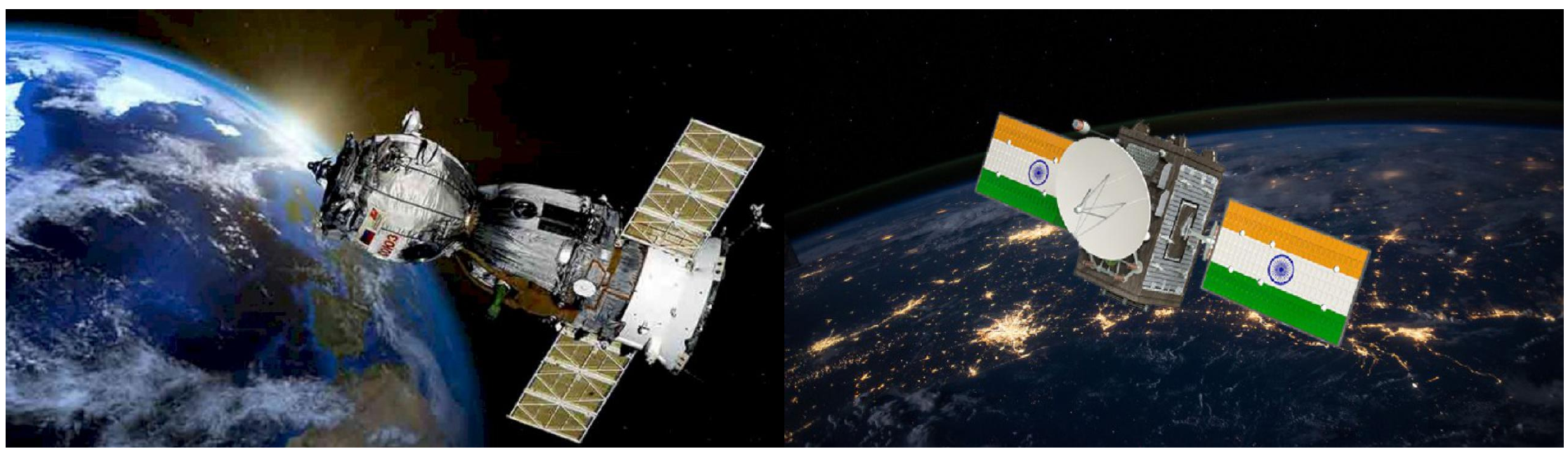}}
    \subfigure[Samples from SPARK dataset. Top-left is the image of ‘Calipso’ satellite with night background of the Earth. Top right is the image of a debris with day background of the Earth. The bottom images are their corresponding depth images \cite{131,132}. Copyright © 2021, IEEE.]{\includegraphics[width=7.5cm,height=6cm]{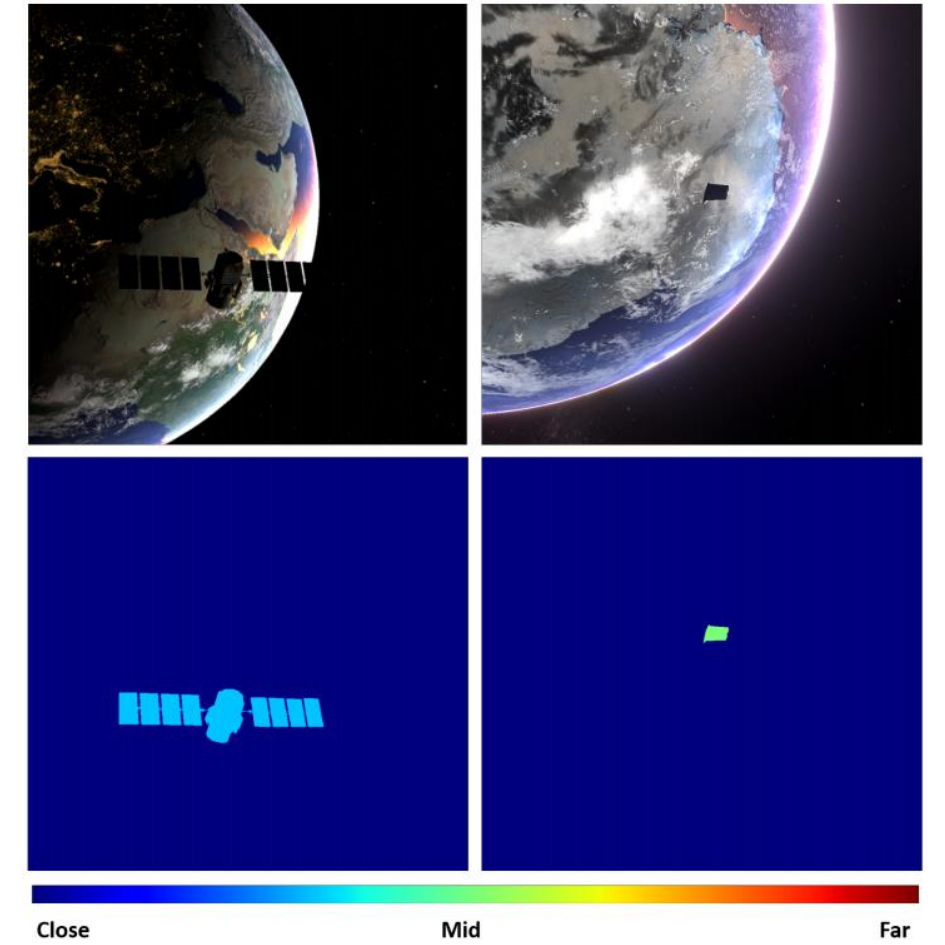}}
    \caption{Example pictures of two datasets for space recognition.}
    \label{fig: 6}
\end{figure}

\subsection{Datasets}
There is no specialized spacecraft datasets for segmentation. The three datasets mentioned above are used for pose estimation and do not provide segmentation annotations. To enrich spacecraft dataset, Hoang \textit{et al.} \cite{133} collect 3,117 images of satellites and space stations, including both synthetic and real images. Examples of collected images are shown in Figure \ref{fig: 6}. To provide a benchmark for this dataset, they also test many object detection and segmentation methods, showing that space-based semantic segmentation is a challenge for models designed on Earth scenarios and is an opening research field. As Space Situational Awareness (SSA) are increasingly emphasized in recent years, the task of space target recognition and detection has been prioritized by some authorities. The competition named \textit{SPAcecraft Recognition leveraging Knowledge of space environment} which was organized by the Interdisciplinary Center for Security, Reliability and Trust (SnT) and the 2021 IEEE International Conference in Image Processing (ICIP), has provided a public dataset \cite{131,132} named SPARK specialized for space target recognition and detection tasks. This dataset is composed of both RGB and depth images, each containing a record-breaking number of 150k. Besides, it has 11 object classes, with 10 classes of spacecrafts and one class of space debris, which is a great complement to the space research field. The samples are shown in Figure \ref{fig: 6}. Different datasets specified in different tasks are listed in Table \ref{tab4} for reference.

\subsection{DL-based fundamental works}
Zeng \textit{et al.} \cite{134} build a nine-layer DCNN (deep convolutional neural network) model, which is a unified framework for feature extraction and classification, to achieve spacecraft recognition. The images they use are rendered by the Systems Tool Kit (STK) \cite{135} and performed with data augmentation before training. Wu \textit{et al.} \cite{136} propose a two-stage CNN-based network for space target recognition. They firstly use a minimum rectangle to cut out all suspected regions, then send the images to a well-trained recognition network to make identification. Chen \textit{et al.} \cite{137} construct an R-CNN based satellite-component-detection model (RSD) to effectively detect satellite components by using optical images. They make some improvements based on Mask R-CNN and achieve pixel-level detection of multiple components. However, sometimes CNN methods can lose attention to the objects in images, leading to misclassification and low accuracy. Therefore, AlDahoul \textit{et al.} \cite{138} propose a decision fusion method to localize and recognize space objects using EfcientNet-v2 and EfcientNet-B4 that were trained on the SPARK dataset.

\subsection{Limitations and solutions}
\textbf{Transfer learning}
Existing spatial non-cooperative object recognition models train isolated models based on specific tasks and datasets without retaining any knowledge that can be transferred from one model to another. While transfer learning can use the knowledge of features and weights in the previously trained model to train the new model, effectively solving the problem of less high-quality data in the recognition task.

For part recognition tasks, due to the change of illumination environment and the interference of starry background and earth background, models trained under ground tests cannot well adapt to the space environment and are prone to failures. Transfer learning can make models quickly learn new knowledge even in an unfamiliar environment, therefore it has great potential in solving multi-domain transfer problems of many space tasks. Gong \textit{et al.} \cite{139} present a transfer learning object detection model based on domain adaptive Faster R-CNN (DA Faster) to detect inclusion and void defects in spacecraft composite structures. Xiang \textit{et al.} \cite{140} introduce transfer learning technology into the field of spacecraft fault diagnosis and successfully solve the problem of "big data and small samples" faced by spacecraft. AlDahoul \textit{et al.} \cite{141} propose a multi-modal learning for spacecraft classification combining vision transformer and end-to-end CNN models based on RGB-D images.

\textbf{Few-shot learning}
Due to the lack of real images of space targets, the model trained with synthetic images can not well adapt to the real space environment. Therefore, few-shot learning can be a tendency to carry out AI-based perception tasks. It is mainly divided into two classes: fast adaptation with meta-learning methods and metric-learning based methods. Fast adaptation with meta-learning methods use a "fine-tuning" strategy to tackle few-shot learning problems, and the training process is hard to implement. While metric-learning based methods learn an information similarity metric which is easier to implement for the training process. Yang \textit{et al.} \cite{142} propose a discriminative deep nearest neighbor neural network (D2N4) to generate discriminative features and improve the recognition accuracy for space targets based on metric-learning. Liu \textit{et al.} \cite{143} incorporate semantic information to extract multiple local visual features and introduce an inner disagreement based domain detection module to solve the bias problem in the generalized zero-shot learning. 


\section{Discussion and Future Development}\label{sec6}

\subsection{Deployment of DL algorithms}
Despite the proliferation of DL-based algorithms for space tasks, their deployment in real-world missions remains a significant challenge. Real-world missions require solutions that can operate within the constraints of latency, inference time, and memory requirements typical of resource-limited space systems. As a result, computationally intensive algorithms are often impractical for real-time, on-board applications.
Currently, only a limited number of deep learning perception missions have been tested and validated on edge computing systems. For instance, a pose estimation software incorporating neural network architectures was developed in \cite{144} to ensure compatibility with Edge Tensor Processing Units (TPU), showcasing the potential of such devices for the next generation of AI-enabled satellites.
Additionally, the use of a hybrid Field Programmable Gate Array (FPGA) and System-on-Chip (SoC) device was explored in \cite{145} to demonstrate efficient on-board inference of CNN-based algorithms, thus verifying the effectiveness of the hybrid architecture.
In another study \cite{146}, causal inference and network quantization techniques were applied to perform 6D pose estimation of space-borne targets. A portion of the quantized network was deployed on a Processing-In-Memory (PIM) FPGA, confirming the feasibility and practical deployment of this approach.
The journey towards the deployability of such DL algorithms in space systems, particularly in resource-constrained environments like those encountered with AI accelerators, is still in its early stages. It is crucial for researchers to focus on developing DL algorithms that not only prove feasible from a theoretical point of view but are also deployable in practical, space-borne contexts.
This imperative highlights the need to address the unique challenges of space missions, including the development of algorithms that are lightweight, efficient, and capable of operating within the stringent constraints of space-based systems. As the field progresses, the goal is to bridge the gap between the promise of DL algorithms and their successful implementation in real-world space missions, thereby unlocking new capabilities and enhancing the performance of space operations.

\subsection{Subsequent tasks based on intelligent perception}
In the context of non-cooperative targets, for service spacecraft engaged in space missions, two pivotal subsequent tasks that are predicated on intelligent visual perception are Guidance, Navigation, and Control (GNC) and the estimation of motion and inertia parameters. Beyond these, a suite of additional subsequent tasks includes consistent tracking and surveillance, threat assessment, and communication and coordination, among others.
Given the extensive research and development efforts dedicated to GNC and parameters estimation, this section will concentrate on these two critical tasks. With precise and reliable perception capabilities, these tasks can be executed effectively, yielding optimal outcomes. This underscores the critical importance of perception in the successful execution of space missions.
The accurate perception of non-cooperative targets is paramount for enabling advanced GNC systems to navigate and control the service spacecraft adeptly. It also facilitates the precise estimation of the motion and inertia parameters of the target, which are essential for planning and executing various mission-critical operations.
Furthermore, the effectiveness of perception extends to other tasks such as maintaining consistent tracking and surveillance of targets, conducting thorough threat assessments to ensure mission safety, and establishing seamless communication and coordination between multiple spacecraft or with ground control.
In essence, the role of perception in space missions is multifaceted and far-reaching. It not only lays the foundation for GNC and parameters estimation but also permeates every aspect of space operations, from tracking and surveillance to communication and beyond. As such, the ongoing advancements in perception technologies are instrumental in pushing the boundaries of what is achievable in space missions, enabling more sophisticated and ambitious endeavors in the realm of space exploration and utilization.

\subsubsection{Autonomous GNC}
Guidance, Navigation and Control (GNC) methodologies are commonly used in various phases of on-orbit missions, especially for a spacecraft-manipulator system. The main steps when performing GNC tasks are: i) Navigation, which refers to the relative state estimation of the targets and environmental interactions. ii) Guidance, which can generate a series of desired states according to the mission requirements and the current states. iii) Control, which inputs the controlling command to the system and performs the guidance instructions \cite{27,147,148,149,150}. 

Intelligent perception technologies are the basis of autonomous GNC tasks, as the navigation process, which is the first step of GNC, needs accurate perception of the observed targets in advance. Moghaddam \textit{et al.} \cite{151} present a comprehensive overview of GNC technologies for space manipulators and help researchers choose appropriate GNC technologies for specific tasks. With the growth of artificial intelligence, AI-based systems show their great advantage of real-time decision-making that is far over conventional spacecraft GNC architectures, which is highly dependent on ground control and human operators. Hao \textit{et al.} \cite{152} introduce two AI modules to replace the corresponding functions in conventional GNC architectures and demonstrate the potential of AI-based technologies in space missions, which is a stepping stone for the future development of autonomous robot systems.

\subsubsection{Motion and inertia parameter estimation}
Motion and inertia parameters of space targets are closely related to their characteristics and dynamics, such as relative position and velocity, attitude quaternion, angular velocity, moment of inertia and inertia tensor. For non-contact vision systems which are usually used in uncooperative situations, Extended Kalman Filter (EKF) based methods are mainstream to perform motion and inertia parameter estimation. Aghili \textit{et al.} \cite{153} present a noise-adaptive Kalman filter for motion estimation and prediction for free-falling tumbling targets, which is computationally efficient and suitable for real-time implementation. Segal \textit{et al.} \cite{154} employ multiple iterated extended Kalman filters (IEKFs) with a maximum a posteriori (MAP) inertia tensor identification algorithm for relative state estimation, which is robust to uncertainties in the inertia tensor. However, their method still involves more power consumption, computational cost. Pesce \textit{et al.} \cite{155} propose a novel method to improve the the relative state estimation with only stereo-vision measurements, comparing the classical EKF with IEKF method during the estimation procedure.

\subsection{Limitations and possibilities}
This part will talk about the limitations of DL-based perception in real applications and explore some possible solutions to improve their feasibility.
\subsubsection{Less independent of Data}
\textbf{Data augmentation}
Data augmentation can generate more training samples for learning-based algorithms, thus effectively tackling the problem of insufficient data, especially in space environment. Due to the scarcity of real-world data, many researchers create synthetic data by 3D rendering engines for more training samples and it has been proved that training on synthetic data can achieve state-of-the-art results \cite{156,157,158}. Generative adversarial network (GAN) \cite{159} is a strong tool for image generation, which is composed of one generator and one discriminator. Through constant training, the generator can create virtual images that are hard to tell by the discriminator. Therefore, data augmentation is also attractive in space tasks.

\textbf{Few-shot/Zero-shot learning}
It is hard to obtain training samples of space targets for learning-based algorithms. Although the spacecraft can stay long-term operation in orbit and transmit much data to ground station, the labeling work is time-consuming and few valid samples are available. Therefore, how to solve the problem of insufficient data has been a key step towards the success of AI-based space technologies. Few-shot learning concentrates on using a small sample of target domains to rapidly generalize to new tasks \cite{160}, which is a promising paradigm for space applications. Zero-shot learning mainly deal with the recognition of classes without labeled samples, the core of which lies in knowledge transfer from seen classes to unseen classes \cite{161}. It has also received much attention in machine learning and computer vision fields, which has great potential in AI-based space applications as well.

\subsubsection{Improved algorithms}
\textbf{Vision Transformers}
Transformers \cite{162,163} have recently gained significant popularity in the field of computer vision and are increasingly outperforming CNNs in a multitude of vision-centric tasks. CNNs possess a remarkable capability for feature extraction; however, they face challenges in capturing the interdependencies among various features. In contrast, the core block of transformer is its self-attention mechanism, which can capture long-distance dependency to fully exploit the relationship. Transformers exhibit the capacity to handle diverse modalities, facilitating cross-domain self-supervised learning. Some transformer-based models with parameter-efficient settings are capable of performing multiple tasks with different modalities, with the quality of strong generalizability. The advantage of parameter-efficient makes it possible to deploy the models on embedded devices with limited memory, especially on space avionics \cite{164}. When dealing with insufficient training data towards general vision intelligence, strong generalizable models \cite{165} can dramatically reduce data demands and facilitate applications in different fields. The models hold vast potential for comprehensive advancement, offering the potential to integrate additional modalities and extend their applicability to a wider array of perceptual tasks.

\textbf{GPT}
Large-scale models have excellent performance with explosive parameters embedded in their networks. 
Here we present the prominent large-scale model known as GPT (Generative Pre-trained Transformers), which has gained widespread popularity across various domains. GPT leverages the transformer architecture to autonomously generate diverse types of content, including images, text, and videos. The release of ChatGPT by OpenAI at the end of 2022 has captured the attention of the global community. This advanced model is capable of delivering responses that are well-aligned with human inquiries and can engage in conversations with a human-like fluency. The model excels as an advanced chatbot, proficiently executing commands provided by operators. Throughout its evolution, the GPT series has progressed through four iterations, from GPT-1 to GPT-4 \cite{166}, consistently enhancing its capabilities to emulate human-like performance. They have shown great potential in achieving artificial general intelligence in the field of AI, assisting humans in a wide range of tasks by generating contents from conversations. For example, producing synthetic spaceborne images at a low cost overcomes the data limitation in space tasks. Based on the data received on Earth, GPTs are capable of generating comprehensive analysis reports, thereby significantly boosting operational efficiency. In conclusion, GPTs have the potential in many applications and play a crucial role in jobs involving processing and understanding natural language. 

\textbf{Diffusion models}
Diffusion model is another generation model, which is used to generate samples similar to training data. The diffusion model operates on the principle of progressively corrupting the training data with Gaussian noise, subsequently reconstructing the original data through the learning of an inverse denoising process. This approach has shown to outperform GANs in the generation of realistic samples, effectively addressing the challenges associated with the unstable adversarial objectives prevalent in GANs. Now diffusion models have shown great pontential in various domains, such as computer vision, data modeling, natural language processing. For instance, the generation of 3D content from natural language inputs using diffusion models can significantly aid researchers in the field of robotics simulation. This is particularly effective when integrated with sophisticated 3D reconstruction technologies such as NeRF \cite{167,168}. From diffusion to ChatGPT, artificial intelligence generated content (AIGC) has gained much attention. Yang \textit{et al.} \cite{169} think that transferring GPT (decoder-only) to diffusion models is interesting for evaluating diffusion-based generation performance. AIGC will showcase fascinating characteristics with the development of diffusion foundation models.

\textbf{Editable NeRF}
NeRF as a novel 3D technology, utilize implicit representations to synthesize high-quality images from different views. A large number of works spring up followed by vanilla NeRF, aiming to achieve faster rendering speed, low power consumption and better visual effects. Nevertheless, the task of editing or modifying objects within the scene has garnered relatively less focus, despite its potential as a pivotal technology within the realm of mixed reality. Editable NeRF \cite{170,171} let users control the scene as they want and maintain the rendering quality at the same time. This technology is of great significance in the future, where human interaction with the scene can make users become more immersive. In space tasks, after 3D reconstruction of the whole scene, we can modify or segment the objects to make a further analysis. 


\textbf{Physics-informed machine learning}
Data driven methods become increasingly popular as machine learning has shown great success in recent years. Many algorithms have been designed with data driven methods for specific tasks. However, when it comes to insufficient data or unsatisfied constraints dictated by natural laws, data-driven methods will reach their limits \cite{172}. Besides, despite easy access of large data nowadays, the function derived based on data has difficult interpretability. Throughout the long history of development, physics researchers have accumulated a substantial body of knowledge concerning the principles and mechanisms of nature. They have formulated explicit physical laws and equations that serve as foundational tools for subsequent generations to build upon and further explore the intricacies of the physical world. Therefore, mechanism models can incorporate the basic principle of the research process and integrate prior knowledge into the learning process, thus improving generalizability, flexibility, and interpretability \cite{173}. There are abundant works based on the combination of data- and knowledge-driven methods, such as physics-informed deep learning \cite{174} and physics-informed neural network (PINN) \cite{175}.

\subsubsection{Multi-source information fusion}
\textbf{Multi-sensor fusion on single agent}
With the development of autonomous vehicles, sensor technology has been emphasized in order to promote the accuracy of environment perception. The demand for autonomous vehicles is growing, and a single sensor on one agent is insufficient to meet these needs. For instance, in the realm of SLAM technology, an integrated approach is often preferred, wherein LiDAR, cameras, and IMUs collaborate to compensate for each other's limitations and maximize their collective operational effectiveness. This multi-sensor fusion approach leverages the strengths of each technology to compensate for individual weaknesses, thereby optimizing the system's overall performance and reliability. According to different coupling methods, there exist loosely and tightly coupled modes, which differentiate with each other in terms of accuracy and calculation \cite{176}. As multi-sensor data fusion can integrate data from different sensors and realize combine advantages, this information-fused technology has attracted wide attention in autonomous driving fields. Multi-sensor fusion techniques \cite{177,178,179} have been investigated by many researchers which can be referred to.

\begin{figure}[H]
\centering
    \includegraphics[width=8cm,height=6cm]{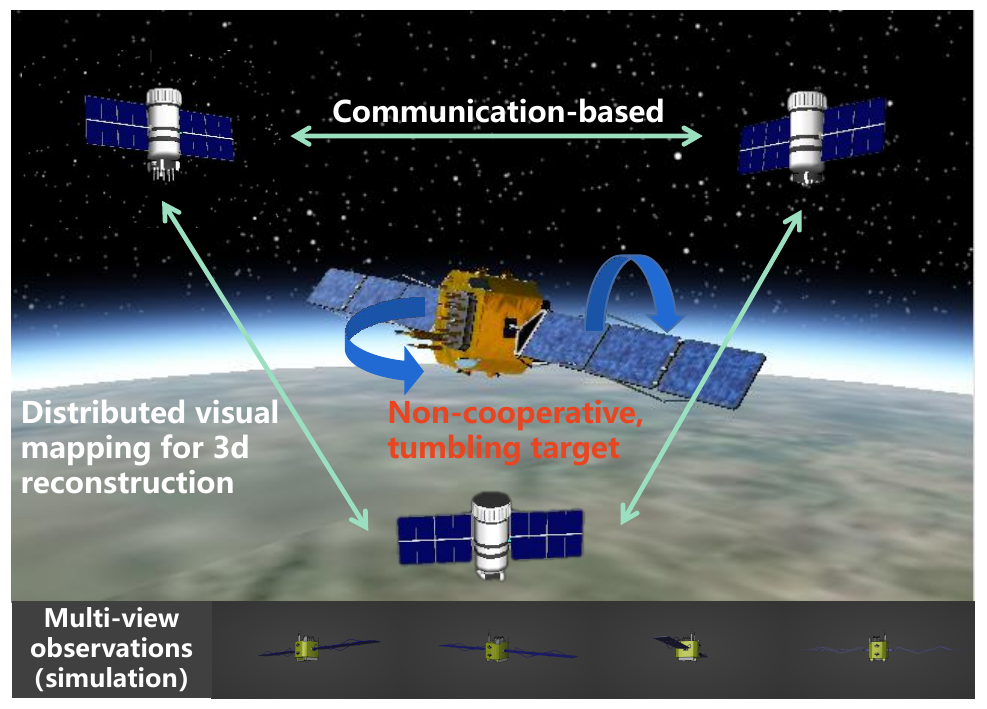}
    \caption{A conceptual illustration of ANS, which cooperate with each other to observe the target.}
    \label{fig: 7}
\end{figure}

\textbf{Multi-agent collaboration}
With the advent of distributed systems in autonomous field \cite{180,181,182}, many explorations for distributed satellite architectures have emerged \cite{148,183}. Small satellites such as Nano and CubeSats have lower cost and power consumption compared with traditional large satellites, which have the potential of being sustainable in the long term. The distribution of functions among multiple agents can reduce the payload and increase redundancy \cite{184}, improving the robustness of a system. Besides, distributed architectures can broaden the observation range and improve the positioning accuracy especially when the observed target is obscured by some unknown space objects. Each satellite of the swarm is equipped with functional sensors and communication avionics, through which they can convey useful information and cooperate to complete a task. The concept of Autonomous Nanosatellite Swarming (ANS) has been widely introduced in recent years, which refers to a group of small spacecraft that operate autonomously and collaboratively \cite{185}. Each spacecraft of ANS is equipped with low size, weight, power and cost (SWaP-C) avionics such as star trackers, short-range cameras, radio-frequency intersatellite links, and chip scale atomic clocks. Figure \ref{fig: 7} is a conceptual illustration of ANS. The distributed architecture are promising as they have the advantage of greater robustness, flexibility, accuracy and sensing coverage. Examples of some distributed space systems include \cite{186,187,188}. Recently, many works \cite{189,190} have focused on asteroid characterization using swarms of nano satellites. To safely approach the target asteroid, the relative state of the spacecraft should be estimated together with the asteroid gravity field, shape, and rotational motion. Multi-agent collaboration has emerged as a burgeoning field of research and is poised to become a key trajectory for the advancement of the space industry. 

\subsubsection{AI chips}
Computing power, as the foundation of Artificial Intelligence (AI), needs to keep pace with the growth of big data and advanced algorithms. Traditional computing architectures can hardly satisfy the computing needs of deep learning. Therefore, AI chips, which refers to chips that are specially designed for AI algorithms, have been a research focus and have a huge room for innovation in various scientific research and applications. High performance with restricted power budget is the demand and expectation for edge computation in various challenging domains such as robotics, underwater, and space. The most prospective AI chips for space applications are Field Programmable Gate Array (FPGA) \cite{191,192}, heterogeneous System-on-Chips (SoC) \cite{193} and neuromorphic processors (NPs) \cite{194}. Compared with CPU and GPU, FPGA has low power consumption, high performance and programmable ability, which is more efficient when dealing with specific tasks. SoC emerges as another promising solution for powerful AI algorithms and computer vision pipelines to carry out computation at the edge. In terms of processors and memories, it has programming flexibility and diversity. NPs represent another efficient opportunity for deploying AI models targeting specific workloads, which can implement spiking neural networks on board power-limited platforms. AI chips have fundamental industrial value and with the speeding up of computing power, AI technology deployed on AI chips can make huge breakthroughs.

\section{Conclusions}
In this review, our objective is to summarize the computer vision tasks for intelligent aerospace perception and to contribute to the growing area of research in this field. Therefore, we introduce three principal visual perception tasks, offering an in-depth survey of their prevailing methodologies and the seminal works that have propelled their evolution. Compared with traditional methods, DL-based visual perception has more advantages such as robustness and preciseness. In the end, we discuss the challenges of DL-based visual perception and shed light on the current cutting-edge techniques in both the computer vision and natural language processing realms. By integrating these state-of-the-art AI-driven technologies, we propose some valuable ideas and promising directions for future space-based research endeavors. Although DL-based algorithms may have too many parameters to be applicable on resource-limited devices, with the rapid development of hardware such as FPGA and custom AI chips, we are confident that the deployment of these sophisticated algorithms on embedded platforms will become feasible in the not-so-distant future.



\Acknowledgements{This work was supported by the National Natural Science Foundation of China (62233005, 62293502), the Programme of Introducing Talents of Discipline to Universities (the 111 Project) under Grant B17017, Fundamental Research Funds for the Central Universities (222202417006) and Shanghai AI Lab.}%

\InterestConflict{The authors declare that they have no conflict of interest.}



\end{multicols}


\begin{thebibliography}{200}

\bibitem{1} Zhao P Y, Liu J G, Wu C C. Survey on research and development of on-orbit active debris removal methods. Science China Technological Sciences, 2020, 63(11): 2188-2210

\bibitem{2} Yang J, Hou X, Liu Y, et al. A two-level scheme for multiobjective multidebris active removal mission planning in low Earth orbits. Science China Information Sciences, 2022, 65(5): 152201

\bibitem{3} Lillie C F. On-orbit assembly and servicing of future space observatories. Space Telescopes and Instrumentation I: Optical, Infrared, and Millimeter. Orlando, Florida, United States: SPIE, 2006. 62652D

\bibitem{4} Ding X L, Wang Y C, Wang Y B, et al. A review of structures, verification, and calibration technologies of space robotic systems for on-orbit servicing. Science China Technological Sciences, 2021, 64(3): 462-480

\bibitem{5} Zhai G, Qiu Y, Liang B, et al. On-orbit capture with flexible tether-net system. Acta Astronautica, 2009, 65(5-6): 613-623

\bibitem{6} Feng F, Tang L N, Xu J F, et al. A review of the end-effector of large space manipulator withcapabilities of misalignment tolerance and soft capture. Science China Technological Sciences, 2016, 59(11): 1621-1638

\bibitem{7} Guariniello C, Delaurentis D A. Maintenance and Recycling in Space: Functional Dependency Analysis of On-Orbit Servicing Satellites Team for Modular Spacecraft. AIAA SPACE 2013 Conference and Exposition. San Diego, CA, 2013. 5327

\bibitem{8} Cui N G, Wang P, GuO J F, et al. Review on the development of space on-orbit service technology. Acta astronautica(in Chinese), 2007, 28(4): 805-811

\bibitem{9} Pan B C, Meng Y H. Relative attitude stability analysis of double satellite formation for gravity field exploration in space debris environment. Scientific Reports, 2023, 13: 1-16

\bibitem{10} Zhang X F, Chen W, Zhu X C, et al. Space advanced technology demonstration satellite. Science China Technological Sciences, 2024, 67(1): 240-258

\bibitem{11} Kaiser C, Sjoeberg F, Delcura J M. Eilertsen: On-Orbit Servicing of a Geostationary Satellite Fleet: OLEV as a Novel Concept for Future Telecommunication Services. 60th IAF Congress, 12.-16.10.2009, Daejeon, Republic of Korea, Paper No.IAC-09.D3.2.4.

\bibitem{12} Kaiser C, Sjoeberg F, Delcura J M, et al. SMART-OLEV--An orbital life extension vehicle for servicing commercial spacecrafts in GEO. Acta Astronautica, 2008, 63(1-4): 400-410

\bibitem{13} Reintsema D, Thaeter J, Rathke A, et al. DEOS–the German robotics approach to secure and de-orbit malfunctioned satellites from low earth orbits. Proceedings of the i-SAIRAS. Japan: Japan Aerospace Exploration Agency (JAXA), 2010: 244-251

\bibitem{14} Wolf T. Deutsche Orbitale Servicing Mission. Space-Administration of the German Aerospace Center. Technical Report, 2011, \url{http://robotics.estec.esa.int/ASTRA/Astra2011/Presentations/Plenary%202/04_wolf.pdf}


\bibitem{15} Zhao C Q, Sun Q Y, Zhang C Z, et al. Monocular Depth Estimation based on Deep Learning: An Overview. Science China Technological Sciences, 2020, 63(9): 1612-1627

\bibitem{16} Tang Y, Zhao C Q, Wang J R, et al. Perception and Navigation in Autonomous Systems in the Era of Learning: A Survey. IEEE Transactions on Neural Networks and Learning Systems, 2023, 34(12): 9604-9624

\bibitem{17} Xia R H, Zhao, C. Q., Zheng, M., et al. CMDA: Cross-Modality Domain Adaptation for Nighttime Semantic Segmentation. In: Proceedings of the IEEE/CVF International Conference on Computer Vision (ICCV). Piscataway, NJ: IEEE, 2023, 21572-21581

\bibitem{18} Zhao C Q, Poggi M, Tosi F, et al. GasMono: Geometry-Aided Self-Supervised Monocular Depth Estimation for Indoor Scenes. In: Proceedings of the IEEE/CVF International Conference on Computer Vision (ICCV). Piscataway, NJ: IEEE, 2023, 16209-16220

\bibitem{19} Liu F C. Application of artificial intelligence in spacecraft. Flight Control and Detection(in Chinese), 2018, 1(1): 16-25

\bibitem{20} Zhang Z, Liu C K, Wang M M, et al. Development and prospects of space intelligent operation. SCIENTIA SINICA Technologica, 2024, 54(2): 289-303

\bibitem{21} Lu K, Liu H, Zeng L, et al. Applications and prospects of artificial intelligence in covert satellite communication: a review. Science China Information Sciences, 2023, 66(2): 121301

\bibitem{22} Hao Y M, Fu S F, Fan X P, et al. Vision Perception Technology for Space Manipulator On-Orbit Service Operations. Unmanned Systems Technology(in Chinese), 2018, 1(1): 54-65

\bibitem{23} Cai H L, Gao Y M, Bing Q J. The research status and key technology analysis of foreign non-cooperative target in space capture system. Journal of Equipment Command Academy(in Chinese), 2010, 20(6): 71-77

\bibitem{24} Rajan K, Saffiotti A. Towards a science of integrated AI and Robotics. Artificial Intelligence, 2017, 247: 1-9

\bibitem{25} Bohg J, Ciocarlie M, Civera J, et al. Big Data on Robotics. Big Data, 2016, 4(4): 195-196

\bibitem{26} Papadopoulos E, Aghili F, Ma O, et al. Robotic manipulation and capture in space: A survey. Frontiers in Robotics and AI, 2021, 8: 686723

\bibitem{27} Kendoul F. Survey of advances in guidance, navigation, and control of unmanned rotorcraft systems. Journal of Field Robotics, 2012, 29(2): 315-378

\bibitem{28} Li C H, Zou H G, Shi D W, et al. Dual-quaternion-based satellite pose estimation and control with event-triggered data transmission. Science China Technological Sciences, 2023, 66(5): 1214-1224

\bibitem{29} Liu M, Liu Q, Zhang L, et al. Adaptive dynamic programming-based fault-tolerant attitude control for flexible spacecraft with limited wireless resources. Science China Information Sciences, 2023, 66(10): 202201

\bibitem{30} Oche P A, Ewa G A, Ibekwe N. Applications and challenges of artificial intelligence in space missions. IEEE Access, 2021, 12: 44481-44509

\bibitem{31} Zhou R, Liu Y, Qi N. Overview of visual pose estimation methods for space missions, Optics and Precision Engineering, 2022, 30(20): 2538-2553

\bibitem{32} Davis T, Baker M T, Belchak T, et al. XSS-10 micro-satellite flight demonstration program. 17th Annual AIAA/USU Conference on Small Satellites, 2003, \url{https://digitalcommons.usu.edu/smallsat/2003/All2003/25/}


\bibitem{33} Debus T, Dougherty S. Overview and Performance of the Front-End Robotics Enabling Near-Term Demonstration (FREND) Robotic Arm. AIAA Infotech@Aerospace Conference and AIAA Unmanned... Unlimited Conference. Seattle, Washington, 2009: 1870

\bibitem{34} Barnhart D, Sullivan B, Hunter R, et al. Phoenix Program Status 2013. AIAA SPACE 2013 Conference and Exposition. San Diego, CA, 2013: 5341

\bibitem{35} Stéphane E, Jürgen T, Lange M, et al. Definition of an Automated Vehicle with Autonomous Fail-Safe Reaction Behavior to Capture and Deorbit Envisat. 7th European Conference on Space Debris. Darmstadt, Germany, 2017, 101

\bibitem{36} Biesbroek R, Innocenti L, Wolahan A, et al. e. Deorbit-ESA’s active debris removal mission. Proc. 7th European Conference on Space Debris, Darmstadt, Germany, 2017, 18-21

\bibitem{37} Caruso B, Mahendrakar T, Nguyen V M, et al. 3d reconstruction of non-cooperative resident space objects using instant ngp-accelerated nerf and d-nerf. arXiv preprint arXiv:2301.09060, 2023

\bibitem{38} Zhang H P, Liu Z Y, Jiang Z G, et al. BUAA-SID 1.0 Space Object Image Dataset. Spacecraft Recovery \& Remote Sensing, 2010, 31(04): 65-71

\bibitem{39} Card M F, Heard Jr W L, Akin D L. Construction and control of large space structures. No. NASA-TM-87689, 1986. 1-20

\bibitem{40} Poirier C, Bataille M, Carazo A R, et al. NASA/GSFC. OSAM-1: On-Orbit Servicing, Assembly, and Manufacturing-1. Retrieved 02 26, 2021, \url{https://www.nasa.gov/mission/on-orbit-servicing-assembly-and-manufacturing-1/}

\bibitem{41} Sedelnikov A V, Salmin V V. Modeling the disturbing effect on the aist small spacecraft based on the measurements data. Scientific Reports, 2022, 12(1): 1300

\bibitem{42} Wang D Y, Hu Q Y, Hu H D, et al. Review of autonomous relative navigation for non-cooperative spacecraft. Control Theory \& Applications(in Chinese), 2018, 35(10): 1392-1404

\bibitem{43} Opromolla R, Fasano G, Rufino G, et al. A review of cooperative and uncooperative spacecraft pose determination techniques for close-proximity operations. Progress in Aerospace Sciences, 2017, 93: 53-72

\bibitem{44} Ruel S, Luu T, Berube A. Space shuttle testing of the TriDAR 3D rendezvous and docking sensor. Journal of Field robotics, 2012, 29(4): 535-553

\bibitem{45} Liu L J, Zhao G P, Bo Y M. Point cloud based relative pose estimation of a satellite in close range. Sensors, 2016, 16(6): 824

\bibitem{46} Preusker F, Scholten F, Matz K D, et al. Topography of vesta from dawn FC stereo images. European Planetary Science Congress 7. San Francisco, USA, 2012

\bibitem{47} Shtark T, Gurfil P. Tracking a non-cooperative target using real-time stereovision-based control: An experimental study. Sensors, 2017, 17(4): 735

\bibitem{48} Segal S, Carmi A, Gurfil P. Vision-based relative state estimation of non-cooperative spacecraft under modeling uncertainty. 2011 Aerospace Conference. Big Sky, MT, USA, 2011. 1-8

\bibitem{49} Feng Q, Liu Y, Zhu Z H, et al. Vision-based relative state estimation for a non-cooperative target. 2018 AIAA Guidance, Navigation, and Control Conference. Kissimmee, Florida, 2018. 2101

\bibitem{50} Fourie D, Tweddle B E, Ulrich S, et al. Flight results of vision-based navigation for autonomous spacecraft inspection of unknown objects. Journal of spacecraft and rockets, 2014, 51(6): 2016-2026

\bibitem{51} Augenstein S. Monocular Pose and Shape Estimation of Moving Targets for Autonomous Rendezvous and Docking. Dissertation for the Master Degree. California: Stanford University, 2011

\bibitem{52} Augenstein S, Rock S M. Improved frame-to-frame pose tracking during vision-only SLAM/SFM with a tumbling target. 2011 IEEE International Conference on Robotics and Automation. Shanghai, China: IEEE, 2011. 3131-3138

\bibitem{53} Deng R, Wang D, E W, et al. Motion Estimation of Non-Cooperative Space Objects Based on Monocular Sequence Images. Applied Sciences, 2022, 12(24): 12625

\bibitem{54} Hao G T, Du X. Advances in optical measurement of position and pose for space non-cooperative target. Laser and Optoelectronics Progress(in Chinese), 2013, 50(8): 240-248

\bibitem{55} Liang B, He Y, Zou Y, et al. Application of time-of-flight camera for relative measurement of non-cooperative target in close range. Journal of Astronautics(in Chinese), 2016, 37(9): 1080

\bibitem{56} Zhang S J, Cao X B, Zhang F, et al. Monocular vision-based iterative pose estimation algorithm from corresponding feature points, Science China Information Sciences, 2010, 53(8): 1682-1696

\bibitem{57} Hu H, Du H, Wang D, et al. Feature-extraction and motion-measurement method for noncooperative space targets. SCIENTIA SINICA Physica, Mechanica \& Astronomica(in Chinese), 2022, 52: 214513

\bibitem{58} Zeng T, Li C X, Liu Q H, et al. Tracking with nonlinear measurement model by coordinate rotation transformation. Science China Technological Sciences, 2014, 57(12): 2396-2406

\bibitem{59} Liang C X, Xue W C, Fang H T, et al. On distributed Kalman filter based state estimation algorithm over a bearings-only sensor network. Science China Technological Sciences, 2023, 66(11): 3174-3185

\bibitem{60} Mo Y, Jiang Z H, Li H, et al. A novel space target-tracking method based on generalized Gaussian distribution for on-orbit maintenance robot in Tiangong-2 space laboratory. Science China Technological Sciences, 2019, 62(6): 1045-1054

\bibitem{61} Ning X, Chen P, Huang Y, et al. Angular velocity estimation using characteristics of star trails obtained by star sensor for spacecraft. Science China Information Sciences, 2021, 64(1): 112209

\bibitem{62} Ruel S, English C, Anctil M, et al. 3DLASSO: Real-time pose estimation from 3D data for autonomous satellite servicing. Proc. ISAIRAS 2005 Conference. Munich, Germany. 2005, 5(8)

\bibitem{63} Blais F, Picard M, Godin G. Accurate 3D acquisition of freely moving objects. 2nd International Symposium on 3D Data Processing, Visualization and Transmission. Thessaloniki, Greece:IEEE, 2004. 422-429

\bibitem{64} Ma, Y. Research on proximity capture technology of failure spacecraft based on slam using lidar. Nanjing: Nanjing University of Aeronautics and Astronautics(in Chinese), 2018. 32-40

\bibitem{65} Cao X B, Zhang S J. An Iterative Method for Vision-based Relative Pose Parameters of RVD Spacecraft. Journal of Harbin Institute of Technology(in Chinese), 2005, 37(8): 1123-1126

\bibitem{66} Opromolla R, Fasano G, Rufino G, et al. Uncooperative pose estimation with a LIDAR-based system. Acta Astronautica, 2015, 110: 287-297

\bibitem{67} Wang K, Liu H, Guo B, et al. A 6D-ICP approach for 3D reconstruction and motion estimate of unknown and non-cooperative target. Chinese Control and Decision Conference. Yinchuan, China, 2016

\bibitem{68} Oumer N W, Kriegel S, Ali H, et al. Appearance learning for 3d pose detection of a satellite at close-range. ISPRS Journal of Photogrammetry and Remote Sensing, 2017, 125: 1-15

\bibitem{69} Shtark T, Gurfil P. Tracking a non-cooperative target using real-time stereovision-based control: An experimental study. Sensors, 2017, 17(4): 735

\bibitem{70} Dor M, Tsiotrasp P. ORB-SLAM applied to spacecraft non-cooperative rendezvous. 2018 Space Flight Mechanics Meeting. Kissimmee, Florida, 2018. 1963

\bibitem{71} Sharma S, D'Amico S. Reduced-dynamics pose estimation for non-cooperative spacecraft rendezvous using monocular vision. 38th AAS Guidance and Control Conference. Breckenridge, Colorado, 2017. 2

\bibitem{72} Mu J Z, Wen K R, Liu Z M. Real-time pose estimation for slow rotation non-cooperative targets. Navigation Position and Timing(in Chinese), 2020, 7(6): 114-120

\bibitem{73} Ge D M, Wang D Y, Zou Y J, et al. Motion and inertial parameter estimation of non-cooperative target on orbit using stereo vision. Advances in Space Research, 2020, 66(6): 1475-1484

\bibitem{74} Peng J, Xu W, Yan L, et al. A Pose Measurement Method of a Space Noncooperative Target Based on Maximum Outer Contour Recognition. IEEE Transactions on Aerospace and Electronic Systems, 2020, 56(1): 512-526

\bibitem{75} Liu K, Wang L, Liu H H, et al. A Relative Pose Estimation Method of Non-Cooperative Space Targets. Journal of Physics: Conference Series, IOP Publishing, 2022, 2228(1): 012029

\bibitem{76} He Y. Modeling and pose measuring of non-cooperative target based on point cloud in close range. Harbin: Harbin Institute of Technology(in Chinese). 2017: 5-12.

\bibitem{77} Li Y F, Wang S C, Yang D F, et al. Aerial relative measurement based on monocular reconstruction of non-cooperation target. Chinese Space Science and Technology(in Chinese), 2016, 36(5): 48-56

\bibitem{78} Dziura M, Wiese T, Harder J. 3D reconstruction in orbital proximity operations. IEEE Aerospace Conference. Big Sky, MT, USA: IEEE, 2017. 1-10

\bibitem{79} Zhang H, Wei Q, Jiang Z. 3D reconstruction of space objects from multi-views by a visible sensor. Sensors, 2017, 17(7): 1689

\bibitem{80} Stacey N, D’Amico S. Autonomous swarming for simultaneous navigation and asteroid characterization. AIAA/AAS Astrodynamics Specialist Conference. Snowbird, UT, 2018, 1

\bibitem{81} Dor M, Tsiotras P. ORB-SLAM applied to spacecraft non-cooperative rendezvous. AAS/AIAA Space Flight Mechanics Meeting. Kissimmee, Florida, 2018. 1963

\bibitem{82} Wong X I, Majji M, Singla P. Photometric stereopsis for 3D reconstruction of space objects. Handbook of Dynamic Data Driven Applications Systems, 2018: 253-291

\bibitem{83} Chen Z S, Zhang C, Su D, et al. 3D reconstruction of spatial non cooperative target based on improved traditional algorithm. In 2021 4th International Conference on Algorithms, Computing and Artificial Intelligence (ACAI). Sanya, China, 2021. 1-6

\bibitem{84} Hu C, Wei M, Huang J, et al. A 3-D shape reconstruction strategy for small solar system bodies with single flyby spaceborne radar. Earth and Space Science, 2023, 10(4): e2022EA002515

\bibitem{85} Zeng F, Yi J, Wang L, et al. Point Cloud 3D Reconstruction of Non-cooperative Object Based on Multi-satellite Collaborations. In 2023 3rd Asia-Pacific Conference on Communications Technology and Computer Science (ACCTCS), 2023, 461-467

\bibitem{86} Dennison K, D’Amico S. Vision-based 3d reconstruction for navigation and characterization of unknown, space-borne targets. Austin, TX, Jan, 2023

\bibitem{87} Moons T, Van Gool L, Vergauwen M. 3D reconstruction from multiple images part 1: Principles. Foundations and Trends® in Computer Graphics and Vision, 2010, 4(4): 287-404

\bibitem{88} Augenstein S. Monocular pose and shape estimation of moving targets for autonomous rendezvous and docking. San Francisco: Stanford University, 2011

\bibitem{89} Takeishi N, Tanimoto A, Yairi T, et al. Evaluation of interest-region detectors and descriptors for automatic landmark tracking on asteroids. Transactions of the Japan Society for Aeronautical and Space Sciences, 2015, 58(1): 45-53

\bibitem{90} Lowe D G. Distinctive Image Features from Scale-Invariant Keypoints. International Journal of Computer Vision, 2004, 60: 91-110

\bibitem{91} Rublee E, Rabaud V, Konolige K, et al. ORB: An efficient alternative to SIFT or SURF. 2011 International Conference on Computer Vision. Barcelona, Spain: IEEE, 2011. 2564-2571

\bibitem{92} Zhou Y, Kuang H Z, Mu J Z. Improved monocular ORB-SLAM for semi-dense 3D reconstruction. Computer Engineering and Applications(in Chinese), 2021, 57(8): 180-184

\bibitem{93} Newcombe R A, Izadi S, Hilliges O, et al. Kinectfusion: Real-time dense surface mapping and tracking. 2011 10th IEEE international symposium on mixed and augmented reality. Basel, Switzerland: IEEE, 2011. 127-136

\bibitem{94} Whelan T, Kaess M, Fallon M, et al. Kintinuous: spatially extended KinectFusion. Robotics \& Autonomous Systems, 2012, 34(4-5): 598-626

\bibitem{95} Whelan T, Leutenegger S, Salas-Moreno R F, et al. ElasticFusion: Dense SLAM without a pose graph. Robotics: science and systems, 2015, 11: 3

\bibitem{96} Newcombe R A, Fox D, Seitz S M. Dynamicfusion: Reconstruction and tracking of non-rigid scenes in real-time. Proceedings of the IEEE Conference on Computer Vision and Pattern Recognition (CVPR). Piscataway, NJ: IEEE, 2015. 343-352

\bibitem{97} Kisantal M, Sharma S, Park T H, et al. Satellite pose estimation challenge: Dataset, competition design, and results. IEEE Transactions on Aerospace and Electronic Systems, 2020, 56(5): 4083-4098

\bibitem{98} Sharma S. Pose estimation of uncooperative spacecraft using monocular vision and deep learning. Stanford University, Department of Aeronautics \& Astronautics, 2019

\bibitem{99} Beierle C R. High fidelity validation of vision-based sensors and algorithms for spaceborne navigation. Stanford University, Department of Aeronautics \& Astronautics, 2019

\bibitem{100} Park T H, Märtens M, Lecuyer G, et al. SPEED+: Next-generation dataset for spacecraft pose estimation across domain gap. 2022 IEEE Aerospace Conference (AERO). Big Sky, MT, USA: IEEE, 2022. 1-15

\bibitem{101} Proença P F, Gao Y. Deep learning for spacecraft pose estimation from photorealistic rendering. 2020 IEEE International Conference on Robotics and Automation (ICRA). Paris, France:IEEE, 2020. 6007-6013

\bibitem{102} Sharma S, D’Amico S. Pose Estimation for Non-Cooperative Rendezvous Using Neural Networks. AAS/AIAA Astrodynamics Specialist Conference. Portland, ME, 2019

\bibitem{103} Park T H, Sharma S, D'Amico S. Towards Robust Learning-Based Pose Estimation of Noncooperative Spacecraft. AAS/AIAA Astrodynamics Specialist Conference. Portland, ME, 2019

\bibitem{104} Chen B, Cao J, Parra A, et al. Satellite pose estimation with deep landmark regression and nonlinear pose refinement. Proceedings of the IEEE/CVF International Conference on Computer Vision Workshops (ICCVW). Piscataway, NJ: IEEE, 2019. 0-0

\bibitem{105} Qiao S, Zhang H, Meng G, et al. Deep-Learning-Based Satellite Relative Pose Estimation Using Monocular Optical Images and 3D Structural Information. Aerospace, 2022, 9(12): 768

\bibitem{106} Gao H, Li Z, Wang N, et al. SU-Net: pose estimation network for non-cooperative spacecraft on-orbit. Scientific Reports, 2023, 13(1): 11780

\bibitem{107} Kelsey J M, Byrne J, Cosgrove M, et al. Vision-based relative pose estimation for autonomous rendezvous and docking. IEEE aerospace conference. Big Sky, MT, USA: IEEE, 2006. 20

\bibitem{108} Xu W, Liang B, Li C, et al. Autonomous rendezvous and robotic capturing of non-cooperative target in space. Robotica, 2010, 28(5): 705-718

\bibitem{109} Žbontar J, LeCun Y. Stereo matching by training a convolutional neural network to compare image patches. Journal of Machine Learning Research, 2016, 17(65): 1-32.

\bibitem{110} Luo W, Schwing A G, Urtasun R. Efficient deep learning for stereo matching. Proceedings of the IEEE Conference on Computer Vision and Pattern Recognition (CVPR). Piscataway, NJ: IEEE, 2016. 5695-5703

\bibitem{111} Seki A, Pollefeys M. Sgm-nets: Semi-global matching with neural networks. Proceedings of the IEEE conference on computer vision and pattern recognition. Honolulu, HI, USA, 2017. 231-240

\bibitem{112} Knobelreiter P, Reinbacher C, Shekhovtsov A, et al. End-to-end training of hybrid CNN-CRF models for stereo. Proceedings of the IEEE conference on computer vision and pattern recognition (CVPR). Piscataway, NJ: IEEE, 2017. 2339-2348

\bibitem{113} Ji M, Gall J, Zheng H, et al. Surfacenet: An end-to-end 3d neural network for multiview stereopsis. Proceedings of the IEEE international conference on computer vision (ICCV). Piscataway, NJ: IEEE, 2017. 2307-2315

\bibitem{114} Kar A, Häne C, Malik J. Learning a multi-view stereo machine. Advances in neural information processing systems, 2017, 30

\bibitem{115} Yao Y, Luo Z, Li S, et al. Mvsnet: Depth inference for unstructured multi-view stereo. Proceedings of the European conference on computer vision (ECCV). Berlin: Springer, 2018. 767-783

\bibitem{116} Chen R, Han S, Xu J, et al. Point-based multi-view stereo network. Proceedings of the IEEE/CVF international conference on computer vision (ICCV). Piscataway, NJ: IEEE, 2019. 1538-1547

\bibitem{117} Xu Q, Tao W. Pvsnet: Pixelwise visibility-aware multi-view stereo network. arXiv preprint arXiv:2007.07714, 2020

\bibitem{118} Xie H, Yao H, Zhang S, et al. Pix2Vox++: Multi-scale context-aware 3D object reconstruction from single and multiple images. International Journal of Computer Vision, 2020, 128(12): 2919-2935

\bibitem{119} Niemeyer M, Mescheder L, Oechsle M, et al. Differentiable volumetric rendering: Learning implicit 3d representations without 3d supervision. Proceedings of the IEEE/CVF conference on computer vision and pattern recognition (CVPR). Piscataway, NJ: IEEE, 2020. 3504-3515

\bibitem{120} Sitzmann V, Zollhöfer M, Wetzstein G. Scene representation networks: Continuous 3d-structure-aware neural scene representations. Advances in Neural Information Processing Systems, 2019, 32

\bibitem{121} Mildenhall B, Srinivasan P P, Tancik M, et al. Nerf: Representing scenes as neural radiance fields for view synthesis. Communications of the ACM, 2021, 65(1): 99-106

\bibitem{122} Müller T, Evans A, Schied C, et al. Instant neural graphics primitives with a multiresolution hash encoding. ACM transactions on graphics (TOG), 2022, 41(4): 1-15

\bibitem{123} Chen Z, Funkhouser T, Hedman P, et al. Mobilenerf: Exploiting the polygon rasterization pipeline for efficient neural field rendering on mobile architectures. Proceedings of the IEEE/CVF Conference on Computer Vision and Pattern Recognition (CVPR). Piscataway, NJ: IEEE, 2023. 16569-16578

\bibitem{124} Cao J, Wang H, Chemerys P, et al. Real-time neural light field on mobile devices. Proceedings of the IEEE/CVF Conference on Computer Vision and Pattern Recognition (CVPR). Piscataway, NJ: IEEE, 2023. 8328-8337

\bibitem{125} Pumarola A, Corona E, Pons-Moll G, et al. D-nerf: Neural radiance fields for dynamic scenes. Proceedings of the IEEE/CVF Conference on Computer Vision and Pattern Recognition. Piscataway, NJ: IEEE, 2021. 10318-10327

\bibitem{126} Song L, Chen A, Li Z, et al. Nerfplayer: A streamable dynamic scene representation with decomposed neural radiance fields. IEEE Transactions on Visualization and Computer Graphics, 2023, 29(5): 2732-2742

\bibitem{127} Mildenhall B, Hedman P, Martin-Brualla R, et al. Nerf in the dark: High dynamic range view synthesis from noisy raw images. Proceedings of the IEEE/CVF conference on computer vision and pattern recognition (CVPR). Piscataway, NJ: IEEE, 2022. 16190-16199

\bibitem{128} Huang X, Zhang Q, Feng Y, et al. Hdr-nerf: High dynamic range neural radiance fields. Proceedings of the IEEE/CVF Conference on Computer Vision and Pattern Recognition (CVPR). Piscataway, NJ: IEEE, 2022. 18398-18408

\bibitem{129} Mergy A, Lecuyer G, Derksen D, et al. Vision-based neural scene representations for spacecraft. 2021 IEEE/CVF Conference on Computer Vision and Pattern Recognition Workshops (CVPRW). Piscataway, NJ: IEEE, 2021. 2002-2011

\bibitem{130} Schwarz K, Liao Y, Niemeyer M, et al. Graf: Generative radiance fields for 3d-aware image synthesis. Advances in Neural Information Processing Systems, 2020, 33: 20154-20166

\bibitem{131} Musallam M A, Gaudilliere V, Ghorbel E, et al. Spacecraft recognition leveraging knowledge of space environment: simulator, dataset, competition design and analysis. 2021 IEEE International Conference on Image Processing Challenges (ICIPC). AK, USA:IEEE, 2021. 11-15

\bibitem{132} Musallam M A, Ismaeil K A, Oyedotun O, et al. SPARK: SPAcecraft Recognition leveraging Knowledge of Space Environment. https://arxiv.org/abs/2104.05978v2

\bibitem{133} Dung H A, Chen B, Chin T J. A spacecraft dataset for detection, segmentation and parts recognition. 2021 IEEE/CVF Conference on Computer Vision and Pattern Recognition Workshops (CVPRW). Nashville, TN, USA: IEEE 2021. 2012-2019

\bibitem{134} Zeng H, Xia Y. Space target recognition based on deep learning. 2017 20th international conference on information fusion (fusion). Xi'an, China: IEEE, 2017. 1-5

\bibitem{135} STK website: \url{https://www.agi.com}

\bibitem{136} Wu T, Yang X, Song B, et al. T-SCNN: A two-stage convolutional neural network for space target recognition. IGARSS 2019-2019 IEEE International Geoscience and Remote Sensing Symposium. Yokohama, Japan: IEEE, 2019. 1334-1337

\bibitem{137} Chen Y, Gao J, Zhang K. R‐CNN‐Based Satellite Components Detection in Optical Images. International Journal of Aerospace Engineering, 2020, 2020(1): 8816187

\bibitem{138} AlDahoul N, Karim H A, De Castro A, et al. Localization and classification of space objects using EfficientDet detector for space situational awareness. Scientific reports, 2022, 12(1): 21896

\bibitem{139} Gong Y, Luo J, Shao H, et al. A transfer learning object detection model for defects detection in X-ray images of spacecraft composite structures. Composite Structures, 2022, 284: 115136

\bibitem{140} Xiang G, Chen W, Peng Y, et al. Deep transfer learning based on convolutional neural networks for intelligent fault diagnosis of spacecraft. 2020 Chinese Automation Congress (CAC). Shanghai, China: IEEE, 2020. 5522-5526

\bibitem{141} AlDahoul N, Karim H A, Momo M A. RGB-D based multi-modal deep learning for spacecraft and debris recognition. Scientific Reports, 2022, 12(1): 3924

\bibitem{142} Yang X, Nan X, Song B. D2N4: A discriminative deep nearest neighbor neural network for few-shot space target recognition. IEEE Transactions on Geoscience and Remote Sensing, 2020, 58(5): 3667-3676

\bibitem{143} Liu B, Dong Q, Hu Z. Semantic-diversity transfer network for generalized zero-shot learning via inner disagreement based OOD detector. Knowledge-Based Systems, 2021, 229: 107337

\bibitem{144} Lotti A, Modenini D, Tortora P, et al. Deep learning for real time satellite pose estimation on low power edge tpu. 2022, arXiv preprint arXiv:2204.03296

\bibitem{145} Cosmas K, Kenichi A. Utilization of fpga for onboard inference of landmark localization in cnn-based spacecraft pose estimation. Aerospace 7, 2020, 159

\bibitem{146} Wang S, Wang S, Jiao B, et al. CA-SpaceNet: Counterfactual Analysis for 6D Pose Estimation in Space. arXiv e-prints, 2022, arXiv:2207.07869

\bibitem{147} Zhou Z, Zhang Z, Wang Y. Distributed coordinated attitude tracking control of a multi-spacecraft system with dynamic leader under communication delays. Scientific Reports, 2022, 12(1): 15048

\bibitem{148} Fazlyab A R, Fani Saberi F, Kabganian M. Fault-tolerant attitude control of the satellite in the presence of simultaneous actuator and sensor faults. Scientific Reports, 2023, 13(1): 20802

\bibitem{149} Yang M F, Liu B, Gong J, et al. Architecture design for reliable and reconfigurable FPGA-based GNC computer for deep space exploration, Science China Technological Sciences, 2016, 59(2): 289-300

\bibitem{150} Xia K, Zou Y. Performance-guaranteed adaptive fault-tolerant tracking control of six-DOF spacecraft. Science China Information Sciences, 2023, 66(1): 119202

\bibitem{151} Moghaddam B M, Chhabra R. On the guidance, navigation and control of in-orbit space robotic missions: A survey and prospective vision. Acta Astronautica, 2021, 184: 70-100

\bibitem{152} Hao Z, Shyam R B A, Rathinam A, et al. Intelligent spacecraft visual GNC architecture with the state-of-the-art AI components for on-orbit manipulation. Frontiers in Robotics and AI, 2021, 8: 639327

\bibitem{153} Aghili F, Parsa K. Motion and parameter estimation of space objects using laser-vision data. Journal of guidance, control, and dynamics, 2009, 32(2): 538-550

\bibitem{154} Segal S, Carmi A, Gurfil P. Vision-based relative state estimation of non-cooperative spacecraft under modeling uncertainty. 2011 Aerospace Conference. Big Sky, MT, USA: IEEE, 2011. 1-8

\bibitem{155} Pesce V, Lavagna M, Bevilacqua R. Stereovision-based pose and inertia estimation of unknown and uncooperative space objects. Advances in Space Research, 2017, 59(1): 236-251

\bibitem{156} Shafaei A, Little J J, Schmidt M. Play and learn: Using video games to train computer vision models. 2016, arXiv:1608.01745

\bibitem{157} Richter S R, Vineet V, Roth S, et al. Playing for data: Ground truth from computer games. Computer Vision-ECCV 2016: 14th European Conference. Amsterdam, The Netherlands: Springer, 2016. 102-118

\bibitem{158} Abu Alhaija H, Mustikovela S K, Mescheder L, et al. Augmented reality meets computer vision: Efficient data generation for urban driving scenes. International Journal of Computer Vision, 2018, 126: 961-972

\bibitem{159} Dewi C, Chen R C, Liu Y T, et al. Synthetic Data generation using DCGAN for improved traffic sign recognition. Neural Computing and Applications, 2022, 34(24): 21465-21480

\bibitem{160} Wang Y, Yao Q, Kwok J T, et al. Generalizing from a few examples: A survey on few-shot learning. ACM computing surveys (csur), 2020, 53(3): 1-34

\bibitem{161} Liu B, Dong Q, Hu Z. Semantic-diversity transfer network for generalized zero-shot learning via inner disagreement based OOD detector. Knowledge-Based Systems, 2021, 229: 107337

\bibitem{162} Dosovitskiy A , Beyer L , Kolesnikov A ,et al.An Image is Worth 16x16 Words: Transformers for Image Recognition at Scale. International Conference on Learning Representations (ICLR). 2021

\bibitem{163} Zhao C, Zhang Y, Poggi M, et al. Monovit: Self-supervised monocular depth estimation with a vision transformer. 2022 international conference on 3D vision (3DV). Prague, Czech Republic: IEEE, 2022. 668-678

\bibitem{164} Likhosherstov V, Arnab A, Choromanski K, et al. Polyvit: Co-training vision transformers on images, videos and audio. Transactions on Machine Learning Research, 2022

\bibitem{165} Shao J, Chen S, Li Y, et al. Intern: A new learning paradigm towards general vision. arXiv preprint arXiv:2111.08687, 2021

\bibitem{166} Wu T, He S, Liu J, et al. A brief overview of ChatGPT: The history, status quo and potential future development. IEEE/CAA Journal of Automatica Sinica, 2023, 10(5): 1122-1136

\bibitem{167} Lin C H, Gao J, Tang L, et al. Magic3d: High-resolution text-to-3d content creation. Proceedings of the IEEE/CVF Conference on Computer Vision and Pattern Recognition (CVPR). Piscataway, NJ: IEEE, 2023. 300-309

\bibitem{168} Poole B, Jain A, Barron J T, et al. Dreamfusion: Text-to-3d using 2d diffusion. arXiv preprint arXiv:2209.14988, 2022

\bibitem{169} Yang L, Zhang Z, Song Y, et al. Diffusion models: A comprehensive survey of methods and applications. ACM Computing Surveys, 2023, 56(4): 1-39

\bibitem{170} Goel R, Sirikonda D, Saini S, et al. Interactive segmentation of radiance fields. Proceedings of the IEEE/CVF Conference on Computer Vision and Pattern Recognition (CVPR). Piscataway, NJ: IEEE, 2023. 4201-4211

\bibitem{171} Yuan Y J, Sun Y T, Lai Y K, et al. Nerf-editing: geometry editing of neural radiance fields. Proceedings of the IEEE/CVF Conference on Computer Vision and Pattern Recognition (CVPR). Piscataway, NJ: IEEE, 2022. 18332-18343

\bibitem{172} Von Rueden L, Mayer S, Beckh K, et al. Informed machine learning–a taxonomy and survey of integrating prior knowledge into learning systems. IEEE Transactions on Knowledge and Data Engineering, 2021, 35(1): 614-633

\bibitem{173} Roscher R, Bohn B, Duarte M F, et al. Explainable machine learning for scientific insights and discoveries. IEEE Access, 2020, 8: 42200-42216

\bibitem{174} Raissi M, Perdikaris P, Karniadakis G E. Physics informed deep learning (part i): Data-driven solutions of nonlinear partial differential equations. arXiv preprint arXiv:1711.10561, 2017

\bibitem{175} Schiassi E, D’Ambrosio A, Scorsoglio A, et al. Class of optimal space guidance problems solved via indirect methods and physics-informed neural networks. Proceedings of 31st AAS/AIAA Space Flight Mechanics Meeting. 2021

\bibitem{176} Xu X, Zhang L, Yang J, et al. A review of multi-sensor fusion slam systems based on 3D LIDAR. Remote Sensing, 2022, 14(12): 2835

\bibitem{177} Aguileta A A, Brena R F, Mayora O, et al. Multi-sensor fusion for activity recognition—A survey. Sensors, 2019, 19(17): 3808

\bibitem{178} Wang Z, Wu Y, Niu Q. Multi-sensor fusion in automated driving: A survey. IEEE Access, 2019, 8: 2847-2868

\bibitem{179} Liang M, Yang B, Chen Y, et al. Multi-task multi-sensor fusion for 3d object detection. Proceedings of the IEEE/CVF conference on computer vision and pattern recognition (CVPR). Piscataway, NJ: IEEE, 2019. 7345-7353

\bibitem{180} Li Z, Tang Y, Fan Y, et al. Formation control of multi-agent systems with constrained mismatched compasses. IEEE Transactions on Network Science and Engineering, 2022, 9(4): 2224-2236

\bibitem{181} Wang J, Hong Y, Wang J, et al. Cooperative and competitive multi-agent systems: From optimization to games. IEEE/CAA Journal of Automatica Sinica, 2022, 9(5): 763-783

\bibitem{182} Hong Y, Jin Y, Tang Y. Rethinking individual global max in cooperative multi-agent reinforcement learning. Advances in neural information processing systems, 2022, 35: 32438-32449

\bibitem{183} Santi G, Corso A J, Garoli D, et al. Swarm of lightsail nanosatellites for Solar System exploration. Scientific Reports, 2023, 13(1): 19583

\bibitem{184} Di Mauro G, Lawn M, Bevilacqua R. Survey on guidance navigation and control requirements for spacecraft formation-flying missions. Journal of Guidance, Control, and Dynamics, 2018, 41(3): 581-602

\bibitem{185} Jin X, Ho D W C, Tang Y. Synchronization of multiple rigid body systems: A survey. Chaos: An Interdisciplinary Journal of Nonlinear Science, 2023, 33(9)

\bibitem{186} Tapley B D, Bettadpur S, Watkins M, et al. The gravity recovery and climate experiment: Mission overview and early results. Geophysical research letters, 2004, 31(9)

\bibitem{187} Krieger G, Moreira A, Fiedler H, et al. TanDEM-X: A satellite formation for high-resolution SAR interferometry. IEEE Transactions on Geoscience and Remote Sensing, 2007, 45(11): 3317-3341

\bibitem{188} Sanchez H, McIntosh D, Cannon H, et al. Starling1: Swarm technology demonstration. 32nd Annual Small Satellite Conference, AIAA/USU, 2018

\bibitem{189} Stacey N, Dennison K, D'Amico S. Autonomous asteroid characterization through nanosatellite swarming. 2022 IEEE Aerospace Conference (AERO). Big Sky, Montana, USA: IEEE, 2022. 1-21

\bibitem{190} Stacey N, D’Amico S. Autonomous swarming for simultaneous navigation and asteroid characterization. AAS/AIAA Astrodynamics Specialist Conference. 2018, 1: 76

\bibitem{191} Cosmas K, Kenichi A. Utilization of FPGA for onboard inference of landmark localization in CNN-based spacecraft pose estimation. Aerospace, 2020, 7(11): 159

\bibitem{192} Giuffrida G, Nannipieri P, Diana L, et al. Satellite Instrument Control Unit with Artificial Intelligence Engine on A Single Chip: ICU4SAT. European Workshop on On-Board Data Processing (OBDP), 2021. 14-17

\bibitem{193} Leon V, Minaidis P, Lentaris G, et al. Accelerating AI and Computer Vision for Satellite Pose Estimation on the Intel Myriad X Embedded SoC. Microprocessors and Microsystems, 2023, 103: 104947

\bibitem{194} Lagunas E, Ortiz F, Eappen G, et al. Performance Evaluation of Neuromorphic Hardware for Onboard Satellite Communication Applications. arXiv preprint arXiv:2401.06911, 2024


















\end{thebibliography}
\end{document}